%% file: main.tex
\documentclass[lettersize,journal]{IEEEtran}
\usepackage{amsmath,amsfonts}
\usepackage{algorithmic}
\usepackage{algorithm}
\usepackage{array}
\usepackage[caption=false,font=normalsize,labelfont=sf,textfont=sf]{subfig}
\usepackage{textcomp}
\usepackage{stfloats}
\usepackage{url}
\usepackage{verbatim}
\usepackage{graphicx}
\usepackage{cite}
\usepackage{graphicx}
\usepackage{booktabs}
\usepackage[numbers]{natbib}
\usepackage[pagebackref]{hyperref}
\usepackage{url}
\usepackage{graphicx}
\usepackage{wasysym} 
\usepackage{xcolor}
\usepackage{mathrsfs}
\usepackage{svg}
\usepackage{array}
\usepackage{makecell}
\usepackage[utf8]{inputenc} % allow utf-8 input
\usepackage[T1]{fontenc}    % use 8-bit T1 fonts
\usepackage{tcolorbox}
\usepackage{url}            % simple URL typesetting
\usepackage{booktabs}       % professional-quality tables
\usepackage{amsfonts}       % blackboard math symbols
\usepackage{nicefrac}       % compact symbols for 1/2, etc.
\usepackage{microtype}      % microtypography
\usepackage{xcolor}         % colors
\usepackage{float}
\usepackage{wrapfig}
\usepackage{caption} 
\usepackage{pifont}
\usepackage{enumitem}
\usepackage{listings}
\usepackage{array}
\usepackage[table]{xcolor}
\usepackage{multirow}
\usepackage{pgf}
\usepackage{booktabs}
\usepackage{colortbl}
\usepackage{tikz}
\usepackage{tabularx}
\usepackage{booktabs}
\definecolor{improvement}{RGB}{34,139,34}
\definecolor{bestcolor}{RGB}{230,240,250}
\usepackage{cleveref}
\begin{document}

\title{RL-RIG: A Generative Spatial Reasoner via Intrinsic Reflection}

\author{Tianyu Wang, Zhiyuan Ma, Qian Wang,  Xinyi Zhang, Xinwei Long, Bowen Zhou,~\IEEEmembership{Fellow,~IEEE} \thanks{Tianyu Wang is with the Zhiyuan College, Shanghai Jiao Tong University (e-mail: wty500@sjtu.edu.cn).}
\thanks{Zhiyuan Ma is with the Huazhong University of Science and Technology (e-mail: {mzyth}@hust.edu.cn).}
\thanks{Qian Wang is with the National University of Singapore.}
\thanks{Xinyi Zhang is with the Shanghai Jiao Tong University.}
\thanks{Xinwei Long is with the Tsinghua University.}
\thanks{Bowen Zhou is with the Tsinghua University \& Shanghai AI Laboratory (e-mail: zhoubowen@tsinghua.edu.cn).}
\thanks{Code is available at https://github.com/wty500/RL-RIG-demo}
}
% <-this % stops a space

\markboth{PREPRINT. }%
{Shell \MakeLowercase{\textit{et al.}}: RL-RIG: A Generative Spatial Reasoner via Intrinsic Reflection}


\maketitle
\input{sec/0_abstract}    
\input{sec/1_intro}
\input{sec/2_related_work}
\input{sec/3_method}
\input{sec/4_experiments}

\input{sec/5_limitations}
\input{sec/6_conclusion}


% \section{References Section}
 
\bibliographystyle{IEEEtran}
\bibliography{IEEEabrv,main}

\newpage
\input{sec/X_suppl}

 

% \end{IEEEbiography}



% \vfill

\end{document}

%% file: sec/0_abstract.tex
\begin{abstract}
  Recent advancements in image generation have achieved impressive results in producing high-quality images. However, existing image generation models still generally struggle with a ``spatial reasoning dilemma'', lacking the ability to accurately capture fine-grained spatial relationships from the prompt and correctly generate scenes with structural integrity. To mitigate this dilemma, we propose \textbf{RL-RIG}, a \textbf{R}einforcement \textbf{L}earning framework for \textbf{R}eflection-based \textbf{I}mage \textbf{G}eneration. Our architecture is comprised of four primary components: Diffuser, Checker, Actor, and Inverse Diffuser, following a \emph{\textbf{``Generate-Reflect-Edit''}} paradigm to spark the Chain of Thought reasoning ability in image generation while addressing the dilemma.  To equip the model with better intuition over generation trajectories, we further develop Reflection-GRPO to train the VLM Actor for edit prompts and the Image Editor for better image quality under a given prompt, respectively. Unlike traditional approaches, which solely produce \textbf{visually stunning yet structurally unreasonable content}, our evaluation metrics prioritize spatial accuracy, utilizing Scene Graph IoU and employing a VLM-as-a-Judge strategy to assess the spatial consistency of generated images on LAION-SG dataset. Experimental results show that RL-RIG outperforms existing state-of-the-art open-source models by up to 11\% in terms of controllable and precise spatial reasoning in image generation.
\end{abstract}
\begin{IEEEkeywords}
Chain of Thought, Vision Language Reasoning,  Image Generation, Image Editing,  Large Language Models.
\end{IEEEkeywords}

\begin{figure*}[htbp]
% \begin{minipage}[b]{.8\linewidth}
\centering
  \begin{tcolorbox}[colback=white, colframe=black, boxrule=0.8pt, sharp corners, boxsep=0pt]
  
  \textbf{ID:} 504620

  \textbf{Prompt:}{\textit{\textbf{\textcolor{blue}{metal lamp}}; \textbf{\textcolor{blue}{adult female person}}; \textbf{\textcolor{blue}{adult male person approaching tall building}}; \textbf{\textcolor{blue}{adult male person walking with adult female person}}; green tree side tall building; modern tram approaching adult male person; St Peter's Square (Manchester) by Trevor Lingard, Local | Manchester | Transport.
  }}
  \end{tcolorbox}
  % \centering
  \setlength{\tabcolsep}{0pt} % 图片之间间隔为0
  \begin{tabular}{@{}c@{\vrule width 1pt}c@{\vrule width 1pt}c@{\vrule width 1pt}c@{}}
    \includegraphics[width=0.23\textwidth]{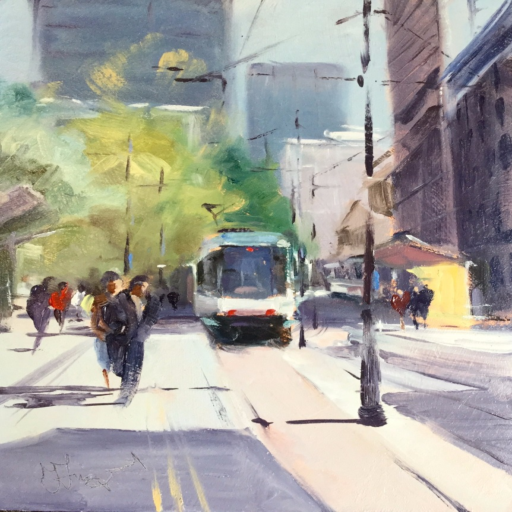} & \includegraphics[width=0.23\textwidth]{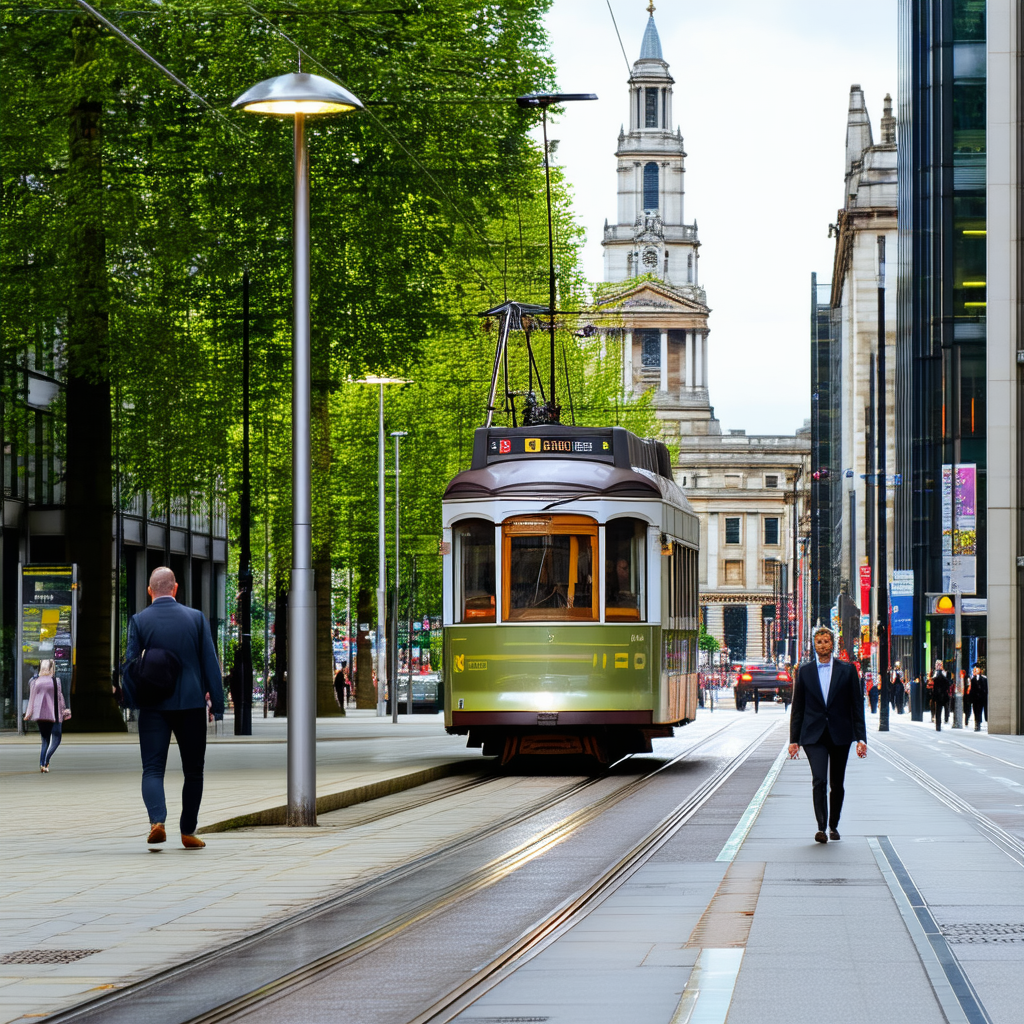} &
    \includegraphics[width=0.23\textwidth]{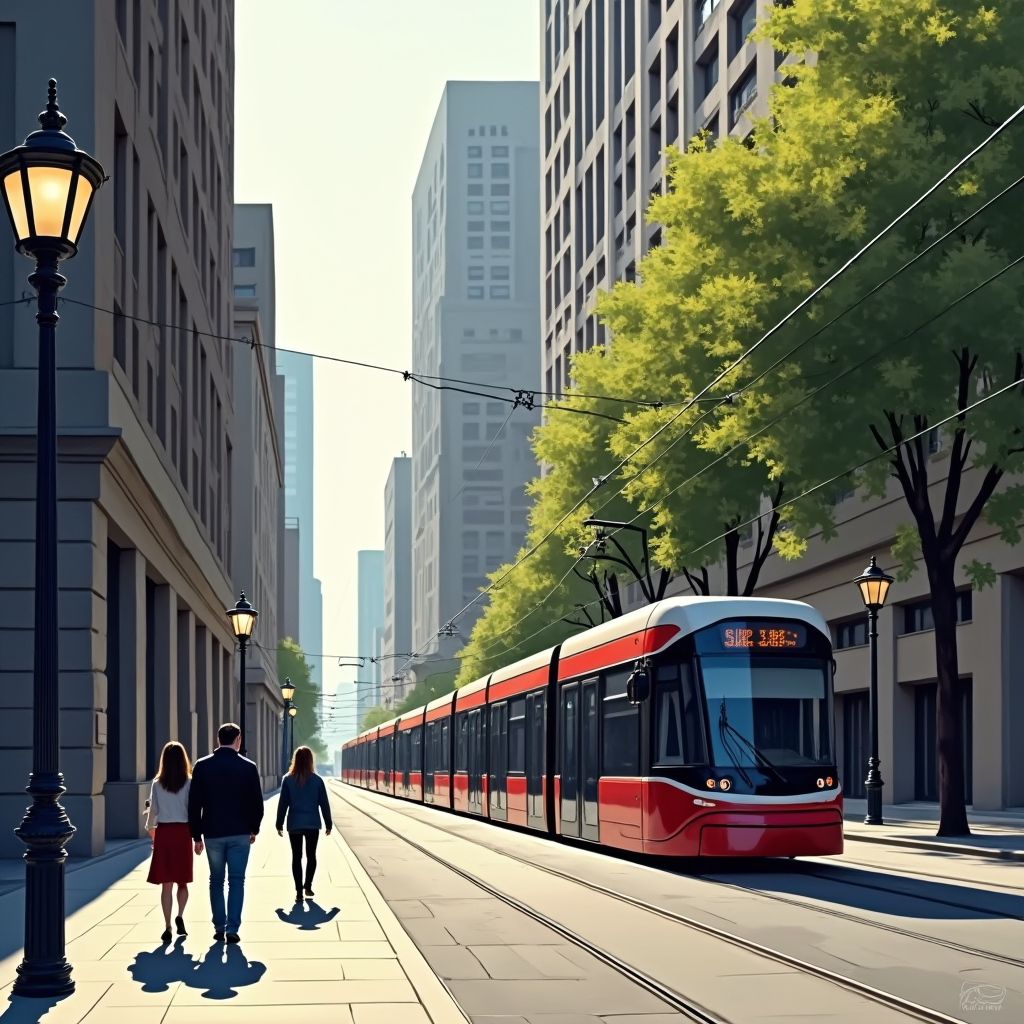} &
    \includegraphics[width=0.23\textwidth]{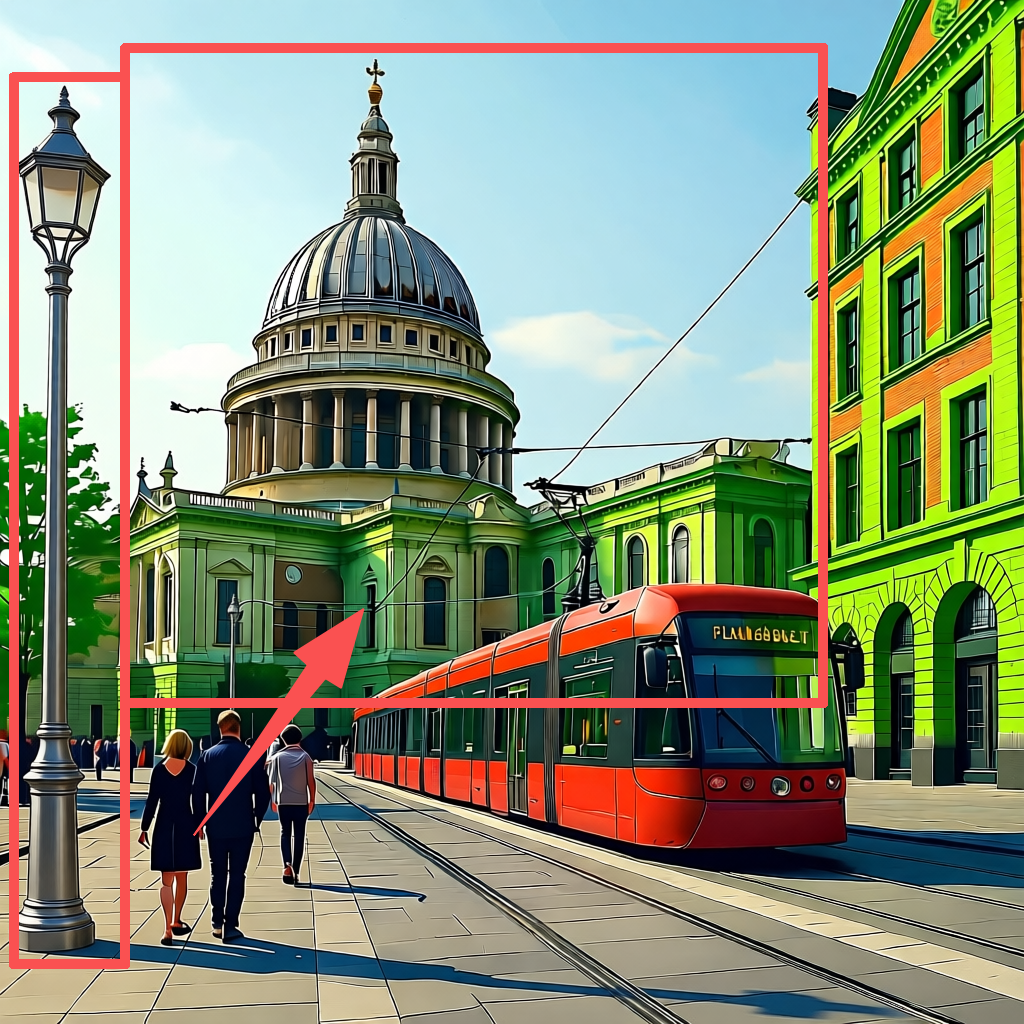} \\
    \textbf{\large GT} & \textbf{\large SD3.5 Large} & \textbf{\large Flux} & \textbf{\large \textcolor{blue}{RL-RIG}}\\
  \end{tabular}

  \begin{tcolorbox}[colback=white, colframe=black, boxrule=0.8pt, sharp corners, boxsep=0pt]
    \textbf{ID:} 523300

  \textbf{Prompt:}{ \textit{female old person walking towards \textbf {\textcolor{blue}{tall building}}; \textbf {\textcolor{blue}{two-wheeled bike}} leaning against tall building; wire basket attached to two-wheeled bike; \textbf {\textcolor{blue}{white fluffy dog sitting in wire basket}}; male young person walking away from tall building; tall building have large sign; \textbf {\textcolor{blue}{female young person riding two-wheeled bike; female young person walking towards tall building}}; Lee-Miller-in-paris-1944}} 
  \end{tcolorbox}
  % \centering
  \setlength{\tabcolsep}{0pt} % 图片之间间隔为0
  \begin{tabular}{@{}c@{\vrule width 1pt}c@{\vrule width 1pt}c@{\vrule width 1pt}c@{}}
    \includegraphics[width=0.23\textwidth]{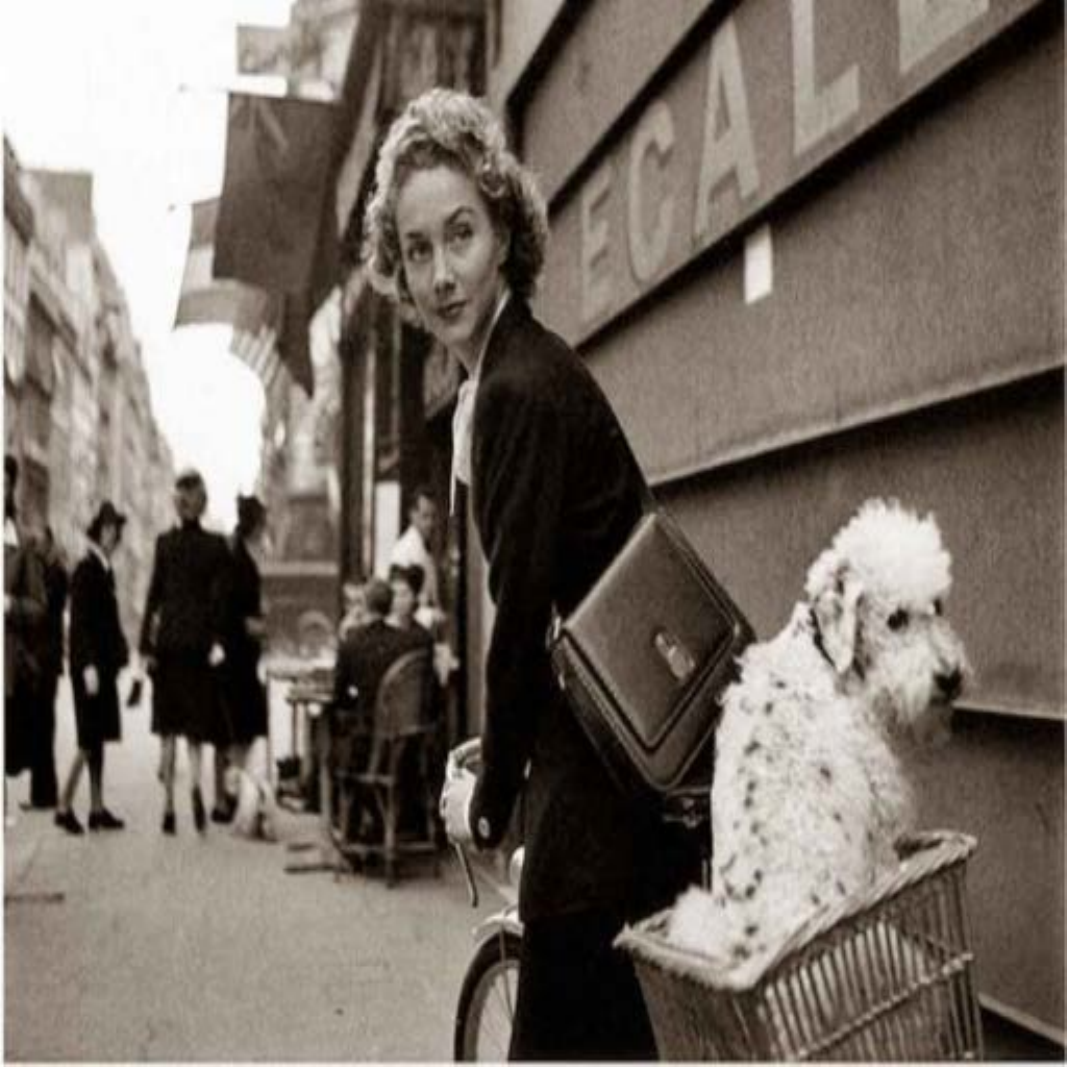}  & \includegraphics[width=0.23\textwidth]{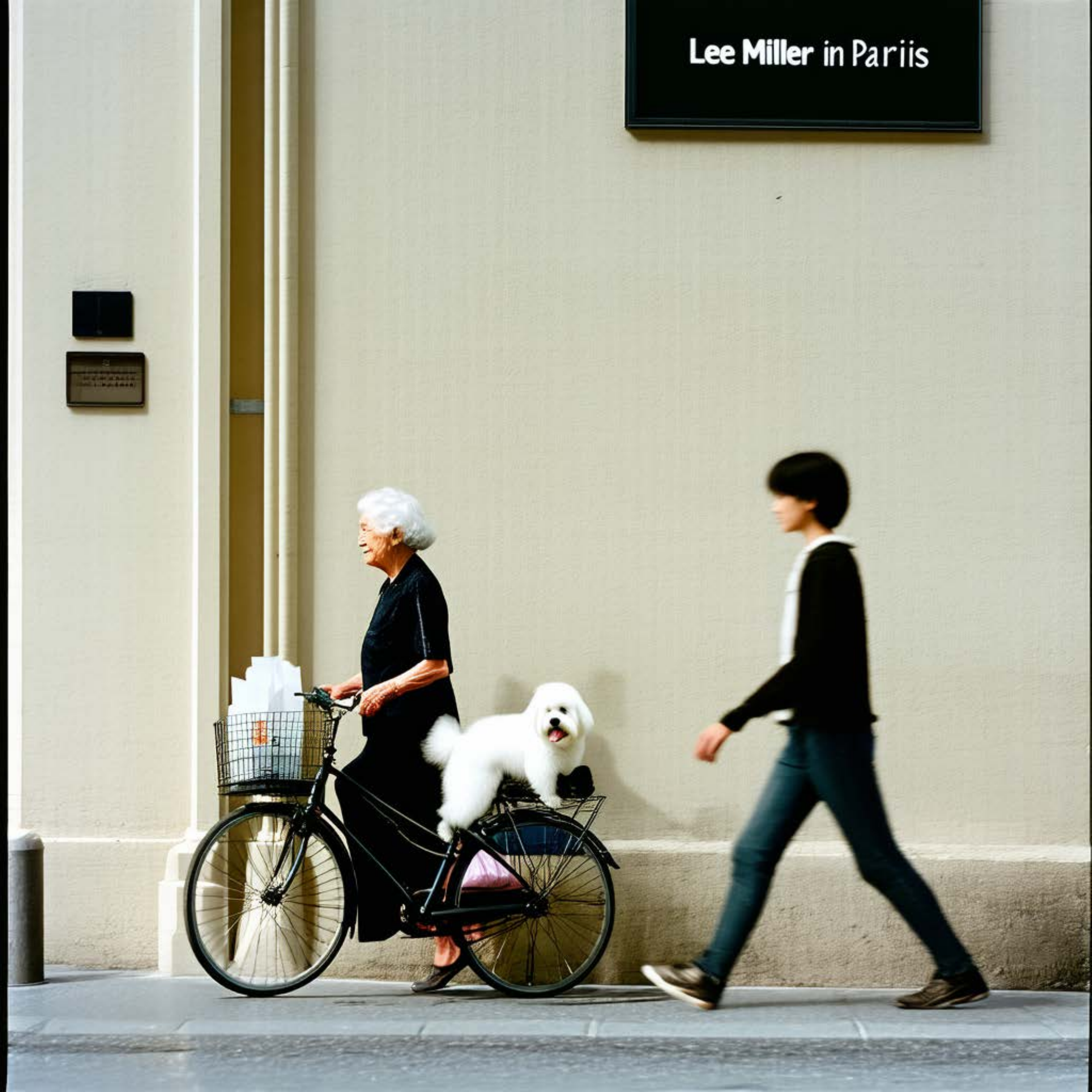} &
    \includegraphics[width=0.23\textwidth]{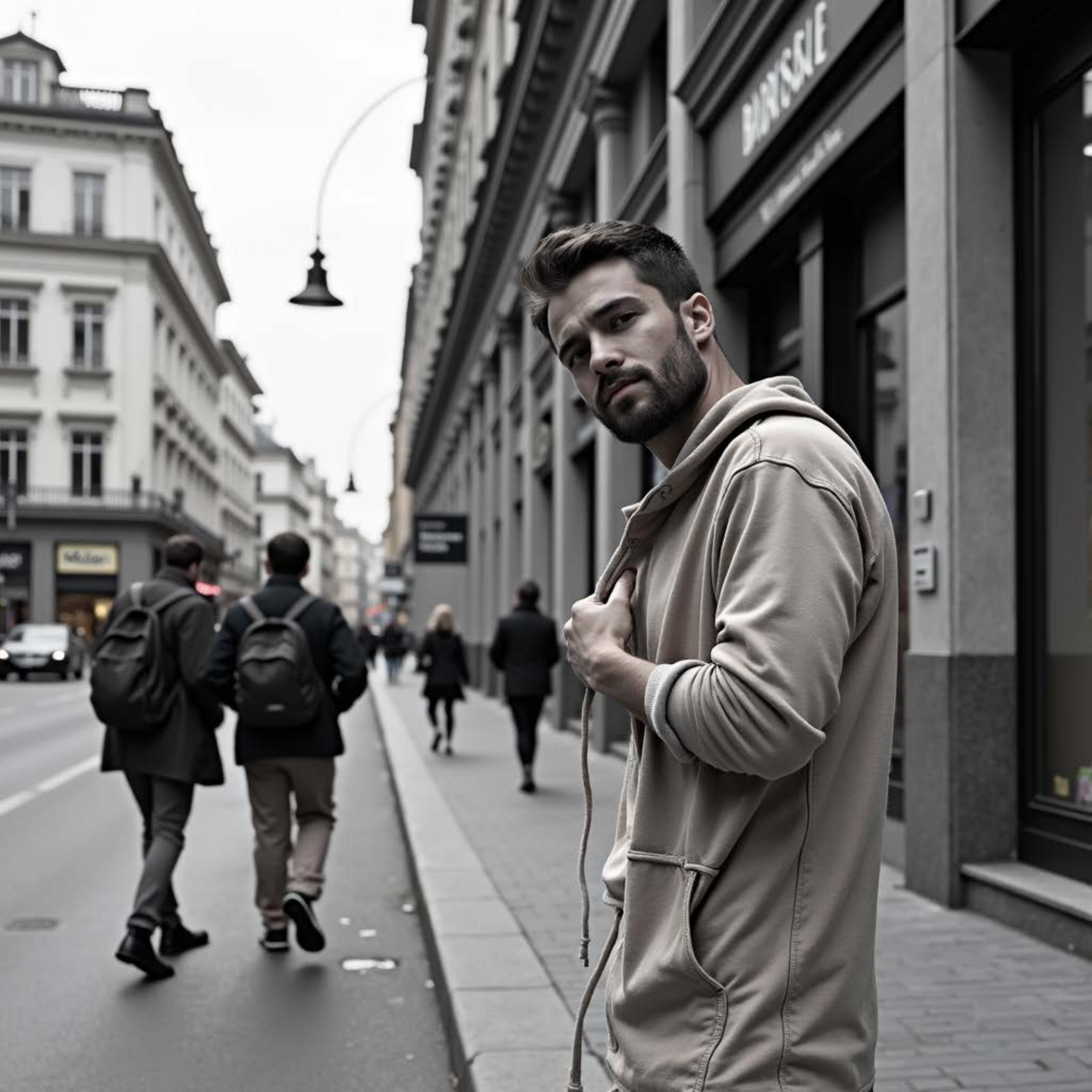} &
    \includegraphics[width=0.23\textwidth]{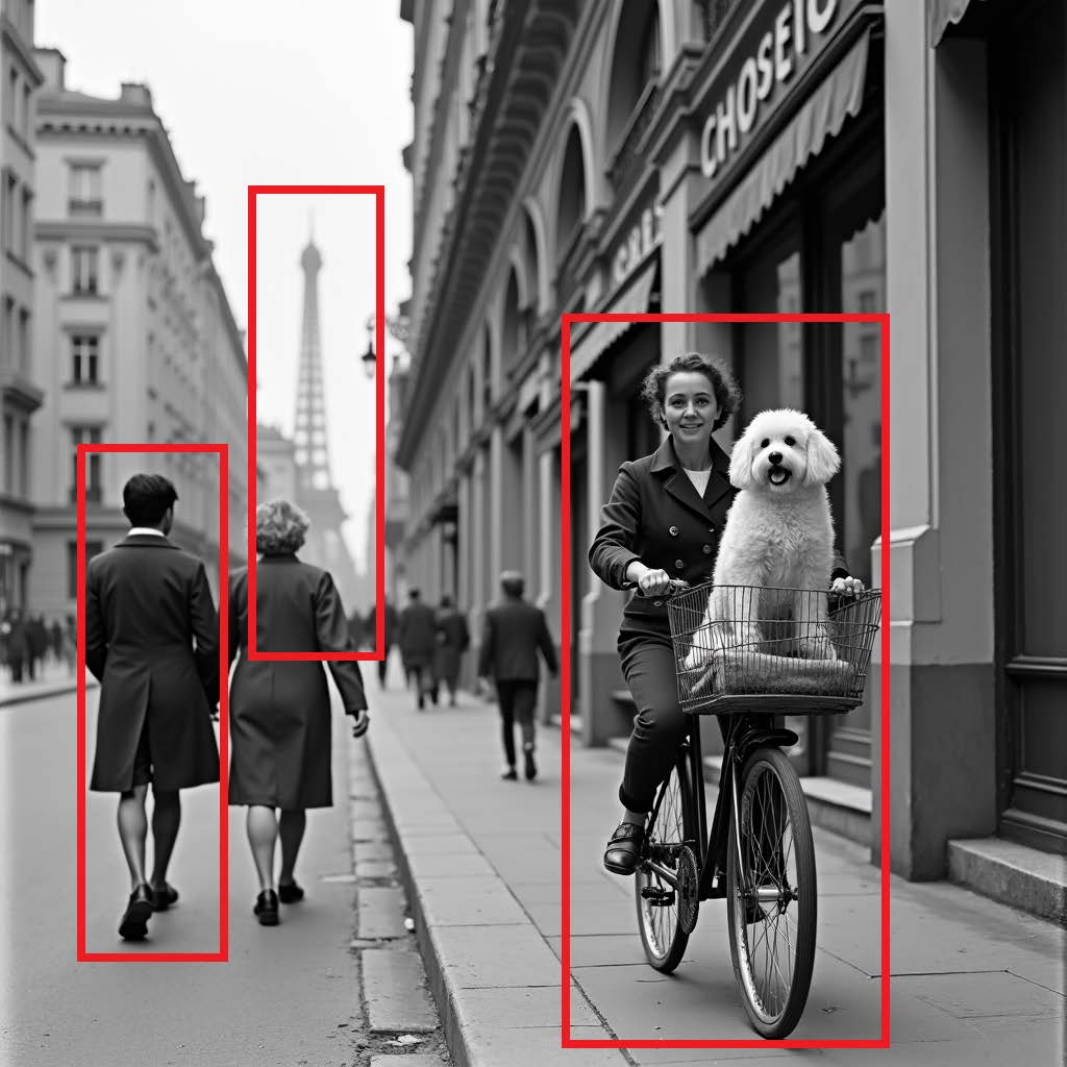}
    \\
    \textbf{\large GT} & \textbf{\large SD3.5 Large} & \textbf{\large Flux} & \textbf{\large \textcolor{blue}{RL-RIG}}\\
  \end{tabular}

   \begin{tcolorbox}[colback=white, colframe=black, boxrule=0.8pt, sharp corners, boxsep=0pt]
    \textbf{ID:} 511475

  \textbf{Prompt:}{ \textit{ \textbf {\textcolor{blue}{tall illuminated lamp}};  \textbf {\textcolor{blue}{tricolored flag on top of large illuminated building}}; golden statue on top corner of large illuminated building; indistinct vehicle in front of large illuminated building;  \textbf {\textcolor{blue}{indistinct person in front of large illuminated building}}; Hanoi Hosts 2nd International Watercolour Painting Exhibition}} 
  \end{tcolorbox}
  % \centering
  \setlength{\tabcolsep}{0pt} % 图片之间间隔为0
  \begin{tabular}{@{}c@{\vrule width 1pt}c@{\vrule width 1pt}c@{\vrule width 1pt}c@{}}
    \includegraphics[width=0.23\textwidth]{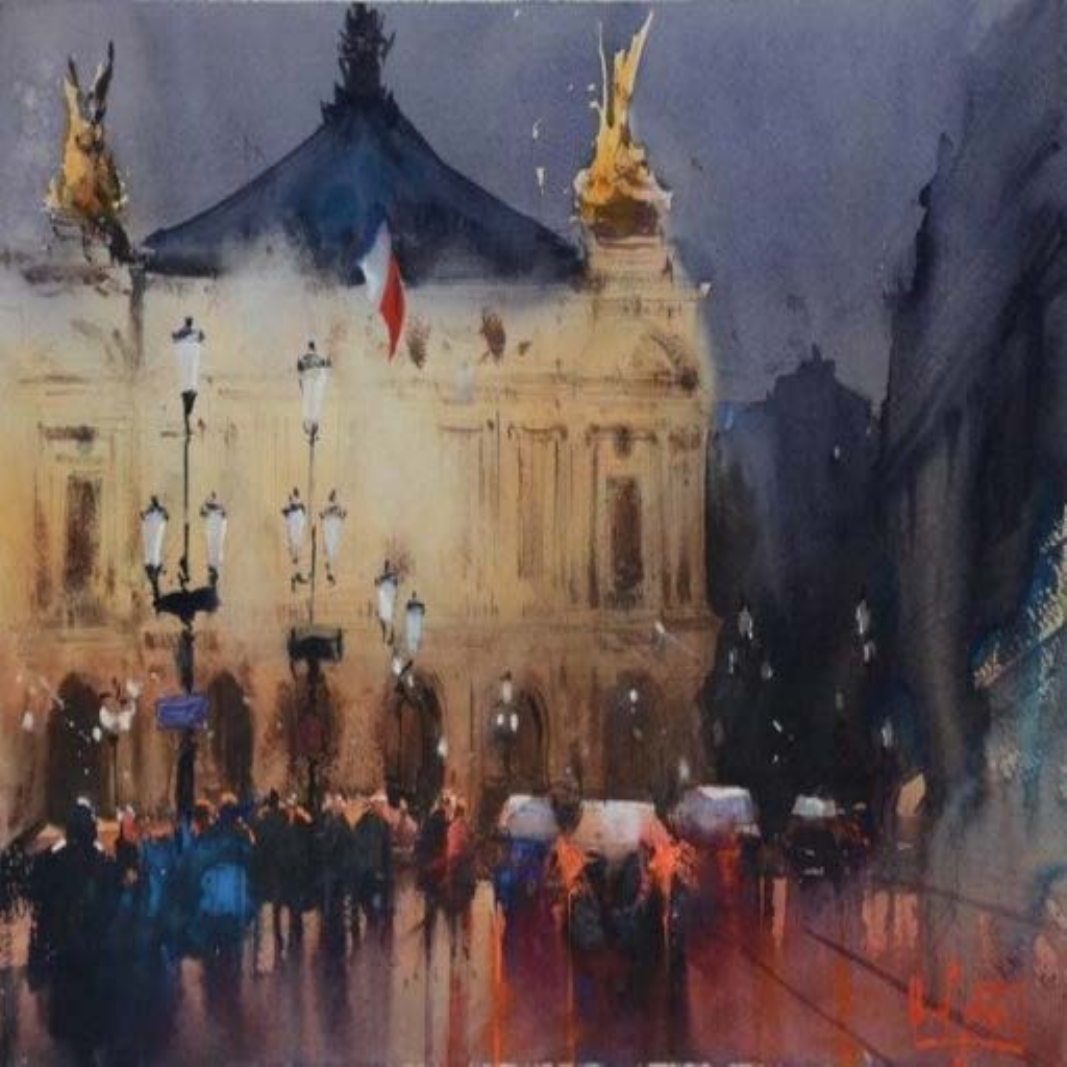} & \includegraphics[width=0.23\textwidth]{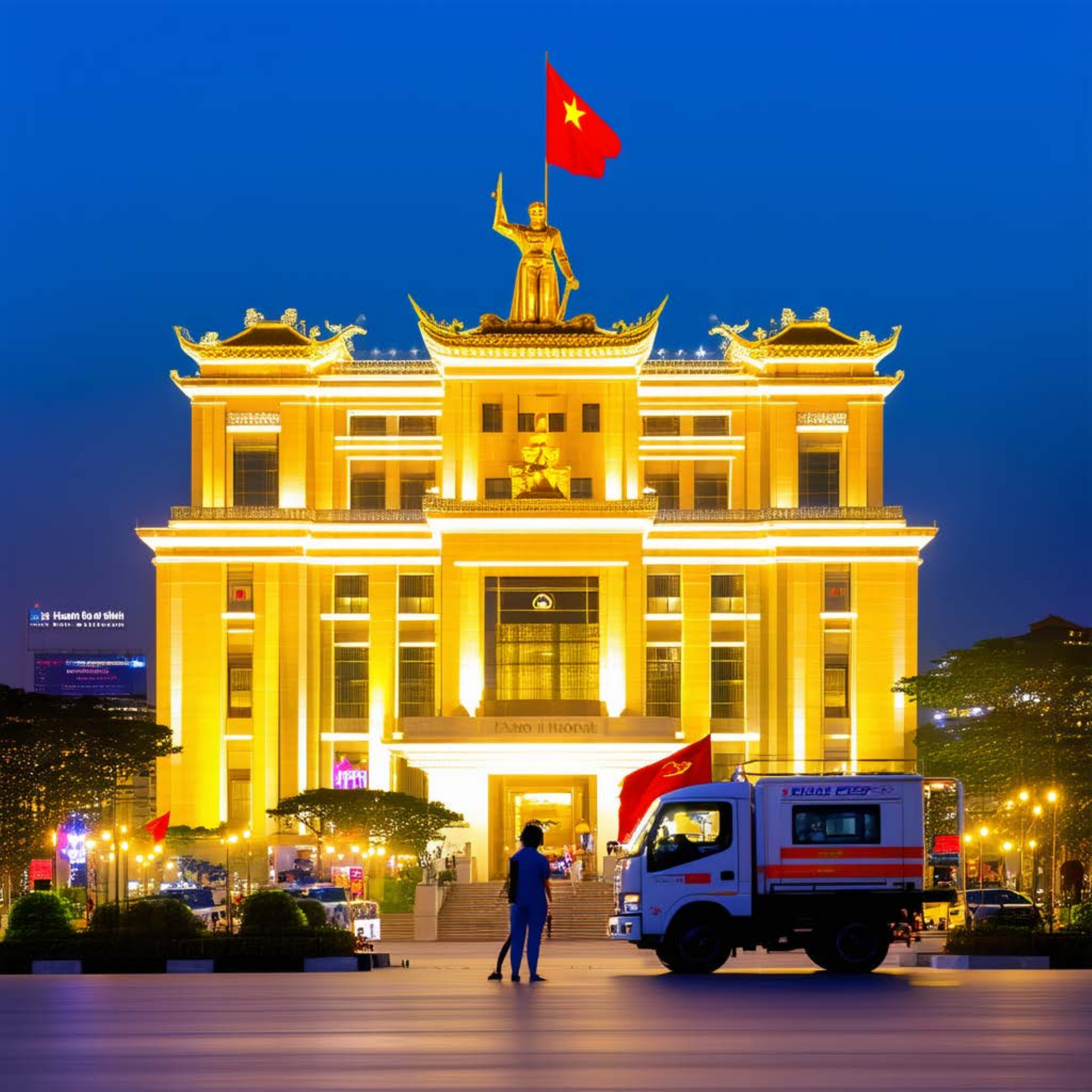} &
    \includegraphics[width=0.23\textwidth]{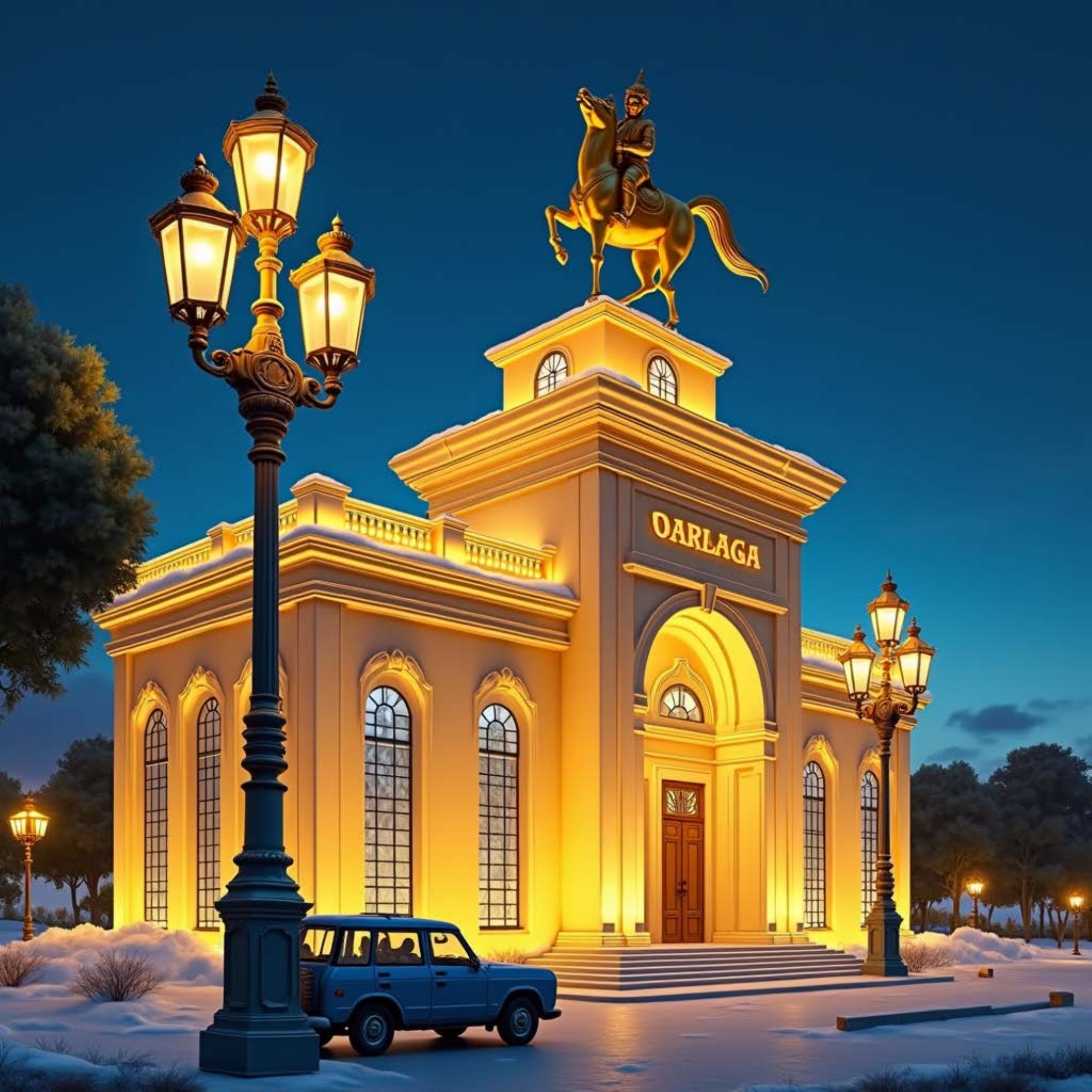} &
    \includegraphics[width=0.23\textwidth]{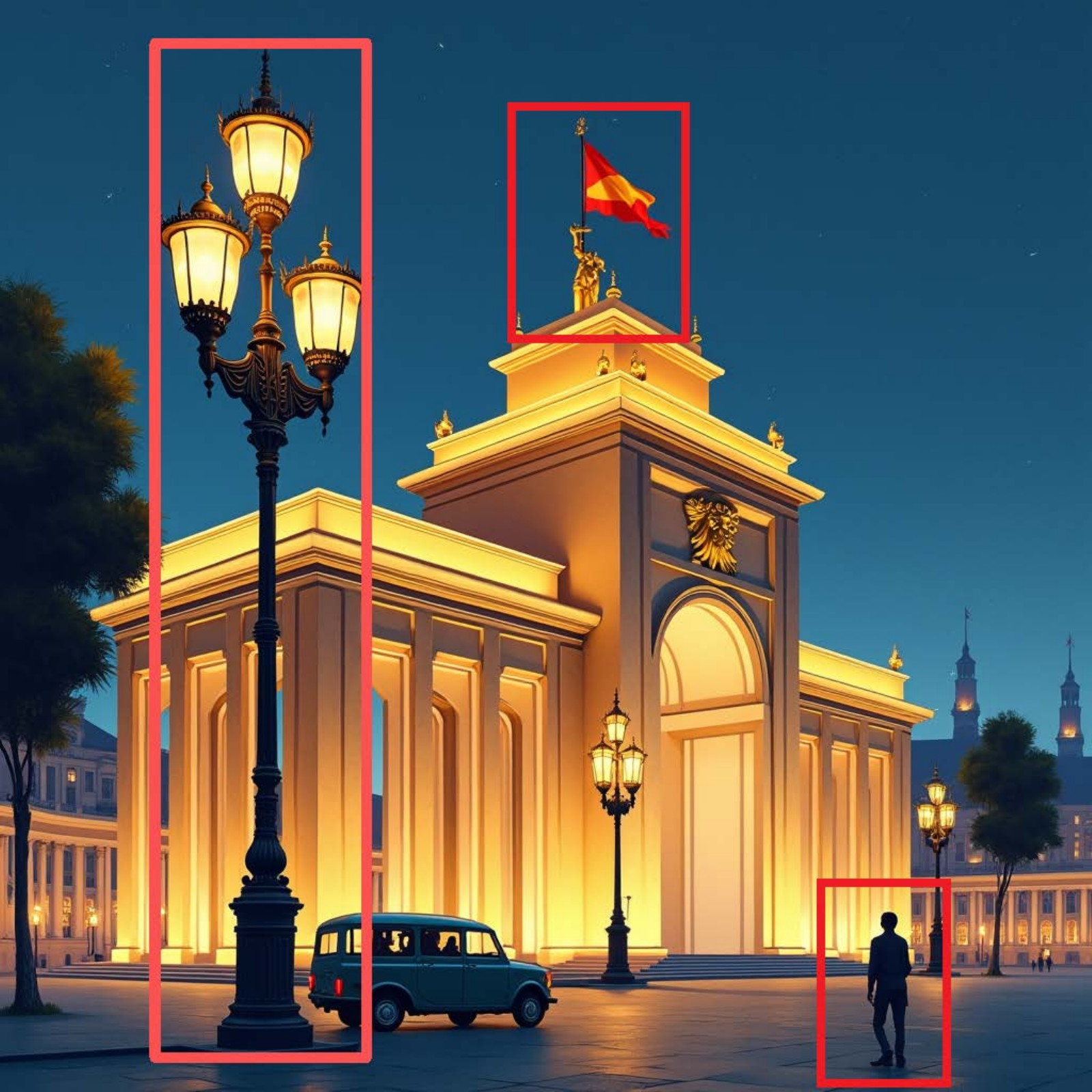}
     \\
    \textbf{\large GT} & \textbf{\large SD3.5 Large} & \textbf{\large Flux} & \textbf{\large \textcolor{blue}{RL-RIG}} \\
  \end{tabular}

  \caption{Comparison of the generated image by Stable Diffusion 3.5 Large, Flux 1.0, RL-RIG \textbf{(choosing Flux as base model)}, and the ground truth image, given a sophisticated prompt with spatial relationships. Our method captures the relations in \emph{\textbf{\textcolor{blue}{blue}}} better, and compensates for the deficiencies in Flux's results. Actually, even the ground truth image does not comply to all the given relationships in the annotation.}
  \label{fig:comparison}
\end{figure*}

%% file: sec/1_intro.tex
\section{Introduction}

\IEEEPARstart{R}{ecent} years have witnessed remarkable progress in text-to-image (T2I) generation such as Stable Diffusion 3.0~\cite{stablediffusion3}, FLUX 1.0 ~\cite{flux2024}, and Janus Pro~\cite{deepseek2025janus}, which can produce high-fidelity and unprecedented images from textual prompts. However, they still face a ``\textit{spatial reasoning dilemma}'', struggling to accurately capture and control fine-grained spatial relationships between objects within generated images while still producing visually impressive images. As illustrated in \autoref{fig:comparison}, while Flux renders high-resolution pictures, it struggles to accurately represent the complicated spatial relationships specified in the prompt. This limitation becomes particularly critical in scenarios that require depicting complex spatial relationships rather than solely pursuing aesthetics \cite{chen2024spatialvlm, song2024hoianimator}.

As two pioneering works, ControlNet~\cite{zhang2023adding} and GLIGEN~\cite{li2023gligen} have been proposed to guide the generation of outcome images. Despite these advancements, they still fall short in achieving flexible and precise control over relative spatial reasoning \cite{lukovnikov2024layout, zhao2023uni, chen2024training}. We attribute the reasons to the following factors: First, in addition to text descriptions,  these methods often require additional user inputs, such as bounding boxes, keypoints, or reference images, which prevents them from realizing an end-to-end scenario. Second, as prompt complexity increases, these methods exhibit limited reasoning ability, making it difficult to accurately interpret all the spatial relationships described. For instance, the CLIP \cite{radford2021learning} encoder in most T2I models supports only at most $77$ input tokens, and its training data primarily focuses on word-level phrases (e.g., \emph{a cat}, \emph{a dog}) instead of global semantics and relational meanings  (e.g., \emph{``a small cute cat stands in front of a yellow dog wagging its tail''}). 

Besides ``spatial reasoning dilemma'', evaluating spatial consistency in generated images is another challenging aspect. Traditional metrics, such as IS \cite{salimans2016improved}, FID \cite{heusel2017gans}, LPIPS \cite{zhang2018unreasonable}, CLIPS score \cite{hessel2021clipscore}, and CMMD \cite{jayasumana2024rethinking}, often focus on the pixel-level distance between generated image and real images, while failing to assess spatial accuracy achieved by the generated image. In other words, images that fulfill the demands of the prompt, even if not similar to the ``ground truth images'', shall also be highly evaluated.

Addressing these limitations involves overcoming several key challenges. First, the complexity of spatial relationships  makes it difficult for text encoders to interpret and represent them all at once. Second, existing image generators and editors, which focus little on spatial layout, tend to produce visually appealing results that lack fidelity to the original prompt. Third, the limited availability of dedicated datasets and evaluation methodologies for complex spatial relationships presents significant challenges for the training and development of frameworks targeting such objectives.

In this work, we propose \textbf{RL-RIG} (a \textbf{R}einforcement \textbf{L}earning framework for \textbf{R}eflection-based \textbf{I}mage \textbf{G}eneration), the first T2I framework to spark models' \textbf{test-time scaling ability} via \textbf{intrinsic reward}.  The core novelty lies in reformulating spatial T2I as a \textbf{trajectory-level reasoning} problem, where generation and editing correspond to exploring different branches. To produce the intrinsic signal for exploration, we provide a \textbf{Generate, Reflect then Edit} framework that can self-evolve in both train time and test time. Instead of merely adding external representations (like bounding boxes), our method addresses the root cause of the \emph{spatial reasoning dilemma}, which is \textbf{lack of explicit verification and reasoning}.
% We reformulate text‑to‑image generation as sampling over \textbf{trajectories}, consisting of noise vectors, inversion states, edit actions, whose branching determines relational structure (\autoref{tree}). Our RL‑RIG activates the intrinsic reward ability for VLM when it acts as a judge, while Reinforcement Learning discourages low‑advantage branches to increase the likelihood of scene‑graph‑consistent endpoints. 

Our key contributions are:
\begin{itemize}
    % \item We recognize the significance of adhering the spatial relationships, which has been ignored by existing metrics. Based on this, we introduce VLM-as-a-Judge and IoU metrics.
    \item We first identify the critical challenge of generating images with complex spatial relationships.  Based on that, our analysis further reveals that current baseline models either rely on supplementary inputs or exhibit constrained reasoning capabilities due to limitations in text encoders.
    \item To cope with the complexity of spatial relationships, we propose RL-RIG, a Self-Reflection based framework to Generate, Reflect and Edit the generated image. We explore test-time scaling and \textbf{a Chain-of-Thought (CoT) reasoning process for complex image generation}, which enables the correctness and rationality of each required spatial relationship one by one.

    \item To optimize the model's intuition reasoning, we explore a two-phase RL strategy featured by post-training the reflection Actor and the Image Editor and reveal such a phenomenon that \textbf{optimizing the generation trajectory can trigger the intrinsic reflection ability of VLMs} to intuitively choose a better trajectory.
    \item To address the lack of criteria for spatial relationships, we utilize Scene Graph IoU \cite{zhang2024multiview} and VLM-as-a-judge \cite{pan2025prometheus} to \textbf{judge the image by faithfulness to the prompt} instead of by the traditional ``ground truth'', showing SOTA performance in GenEval and SG-IoU benchmark.
\end{itemize}
\vspace{-1em}

%% file: sec/2_related_work.tex
\section{Related Works}

Recent work on spatially controllable image generation has focused on incorporating explicit layout or grounding inputs into diffusion models. Works like GLIGEN \cite{li2023gligen}, ControlNet \cite{zhang2023adding}, layout-diffusion \cite{zheng2023layoutdiffusion} and FreestyleNet \cite{xue2023freestyle} have demonstrated effective ways to guide image generation based on specific spatial inputs. More recent approaches have extended these capabilities by adding style control \cite{wang2025stay}.  Beyond bounding-box layouts, there are also scene-graph approaches \cite{farshad2023scenegenie, wang2025scene}.  However, most of these techniques require supplementary information beyond text prompts, which limits their practical applicability in end-to-end scenarios.

Another thread of work applies Reinforcement Learning (RL) and preference optimization to steer diffusion models toward better alignment or specific objectives, such as DDPO \cite{black2023training}, Diffusion-DPO \cite{wallace2024diffusion}, and Diffusion-RPO \cite{gu2024diffusion}. These works illustrate that policy-gradient methods (PPO \cite{schulman2017proximal}, DPO \cite{rafailov2023direct}, RPO \cite{yin2024relative}, etc.) can be used to directly optimize diffusion models for complex rewards, analogously to RLHF \cite{ouyang2022training} in vision language models \cite{pan2025metaspatial}. Nevertheless, few studies apply group-reward based preference optimization, like GRPO \cite{guo2025deepseek}, into image generation models.

A related set of methods uses Vision Language Models or multi-step reasoning to improve control and alignment \cite{wen2023improving,li2025reflect,sun2025marmot, jiang2025t2i}. There has also been progress on rectified-flow generative models and diffusion inversion for more precise control and editing \cite{wang2024taming, patel2024steering, rout2024semantic}, although there are few approaches to leverage the reasoning ability of VLMs for better generative results.

Our work bridges these research directions by combining reinforcement learning with VLM-based reasoning in an intrinsic reflection framework that addresses the spatial reasoning challenges faced by current image generation systems, only requiring plain text input.

\section{Preliminaries}

We focus on difficult text-to-image generation challenges where the input text is extremely sophisticated with many spatial relationships or other requirements, and we have no additional inputs like bounding boxes. We formalize our problem as follows: let $Q=\{Q_1,Q_2,...,Q_n\}$ be the set of all given text requirements, and $O=\{o_1,o_2,...,o_m\}$ be the set of all objects with specific features. Each text requirement $Q_i$ can describe either a new object $o_i$ (with or without adjectives), a spatial relationship $(o_{x_i},r_i,o_{y_i})$ between given objects, meaning that the object $o_{x_i}$ has a relationship $r_i$ over $o_{y_i}$,  or a plain-text description  $d_i$  for other properties of the image. Formally,
\begin{equation}
    Q_i=\begin{cases}
        o_i,\quad &\text{$Q_i$  is an object}\\
        (o_{x_i},r_i,o_{y_i}), \quad &\text{$Q_i$  is  a relationship}\\
        d_i,\quad &\text{otherwise}
    \end{cases}
\end{equation}
The Image Generator $\mathcal{G}(Q)=f_{\theta,\xi}(x_T,T)=I$ takes in only the text prompts, and outputs an corresponding image subject to the requirements in the text. We suppose its output depends on the network parameter $\theta$ and the random seed $\xi$. 

The Image Editor $\mathcal{E}(I, Q)=I^\prime$ takes in an image and an edit prompt, and outputs the edited image. We suppose it is an inversion-based model, which contains inversion stage $f^{-1}_{\theta,\xi^\prime}(x_t,t|\Phi)$ and reversion stage $f_{\theta,\xi^\prime}(x_t^\prime,t|\Phi)$. Here, $t$ is the number of inversion steps, and $\Phi$ is the text embedding of $Q$.

To evaluate whether the requirements are all satisfied, we can employ a checker:
\begin{equation}
    \mathcal{C}(I, Q)= \frac{|\hat{Q}|}{|Q|}
    \label{checker}
\end{equation}
where $\hat{Q}\subseteq Q$ is the set of all satisfied requirements, and $|\cdot|$ means the number of requirements in the given set. In our definition, the checker outputs a score, which represents the proportion of satisfied requirements. Although the checker's output can influence the reward function, we do not utilize the concept of ``Critic'' as in Reinforcement Learning, because the Temporal Difference (TD) is hard to compute and propagate in this scenario.

Let $\tau = \{x_T,\ldots,x_0\} \cup \{a_1,\ldots,a_K\}$ denote a generation trajectory composed of latent states and edit actions. The Generator and Editor induce a stochastic transition $P(\tau)$. After determining $\tau$, $I$ is determined accordingly by the generation and editing process. Our objective is to maximize the expected scene‑graph fidelity: 
 \begin{equation}
 \max~ \sum_{Q} \mathbb{E}_{\tau\sim P}[ \mathcal{C}(\tau, Q)].
 \end{equation}
We employ other related components to rewrite this goal into a component-related form. Let $\mathcal{VLM}(I,  Q)=Q^\prime$ to be  a Vision Language Model that takes in an image and text prompts, and output text responses. Here, we treat the set of requirements as plain text.

We can then reformulate the objective of spatial image generation to be producing a multi-agent collaborative framework $\mathcal{F}_{\mathcal{G},\mathcal{E},\mathcal{C},\mathcal{VLM},\theta}(Q)=I$ with trainable parameters $\theta$ that satisfies as many requirements as possible:
\begin{equation}
    \max~ \sum_{Q} \mathbb{E}[\mathcal{C}(\mathcal{F}_{\mathcal{G},\mathcal{E},\mathcal{C},\mathcal{VLM},\theta}(Q),Q)].
\end{equation}

%% file: sec/3_method.tex
\section{Method}
% \begin{wrapfigure}{l}{0.5\textwidth}
%     \vspace{-1em}
%   \includesvg[width=\linewidth]{img/fig2.svg}
%   \caption{The overview of RL-RIG. Here, we suppose the Image Editor is composed of an inverse diffuser and a diffuser.}
%     \label{overview}
%     \vspace{-2em}
% \end{wrapfigure}
\begin{figure*}[htbp]
    \centering
    \includegraphics[width=\linewidth]{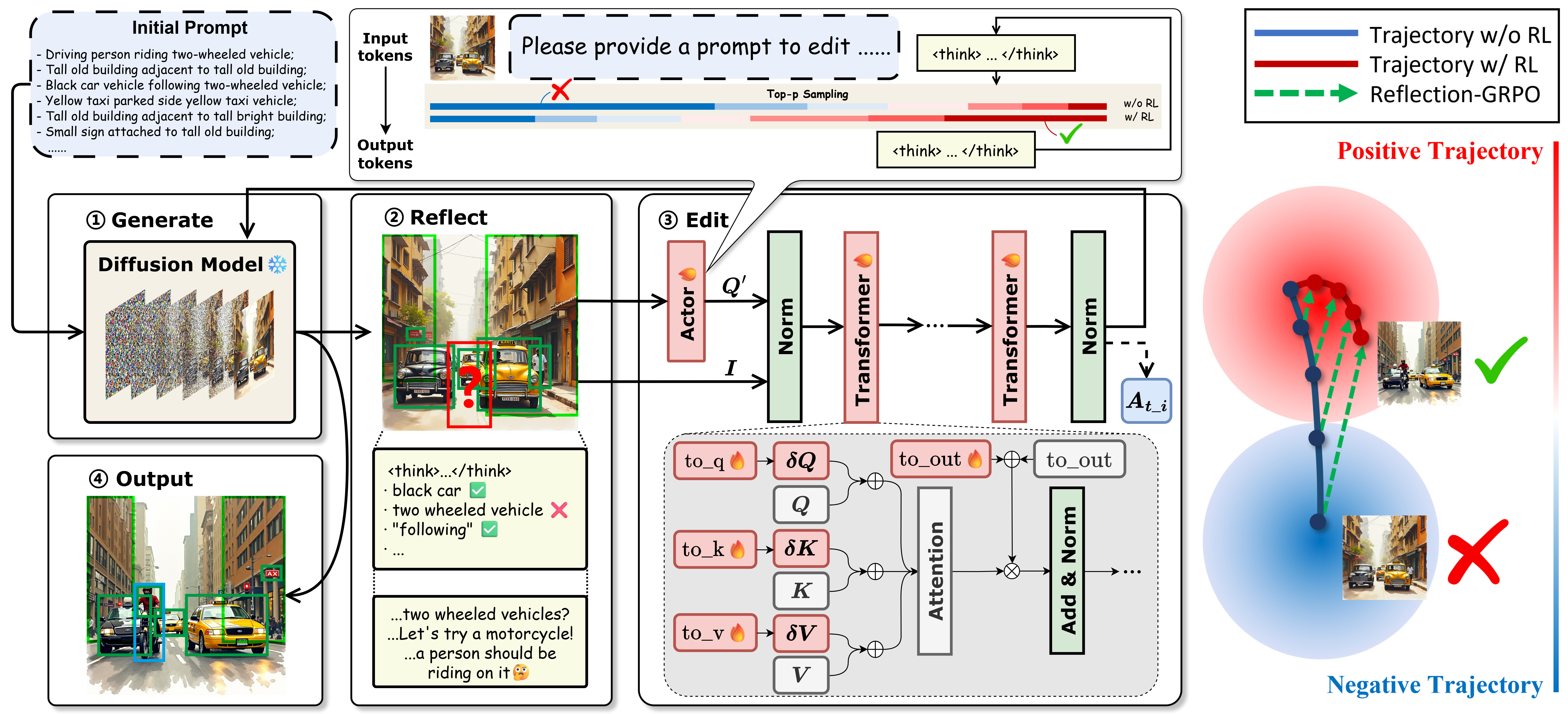}
    \caption{The overview of RL-RIG. The generation phase abides a Generate-Reflect-Edit paradigm; the training phase aims at shifting trajectories, assigning greater probabilities for positive trajectories while discouraging negative trajectories. }
    \label{overview}
\end{figure*}

\subsection{Overview} 

% Since current image generation models struggle to process complex spatial relationships all at one time, we devise a \textbf{Generate-Reflect-Edit} framework that leverages multi-agent Vision Language Models (VLMs) to iteratively improve spatial reasoning in image generation, shown in \autoref{tree} and \autoref{overview}. Our RL-RIG framework addresses spatial reasoning limitations through iterative refinement.
As shown in \autoref{overview}, given a complex prompt with multiple spatial constraints, we devise RL-RIG, a \textbf{Generate-Reflect-Edit} framework that: \textbf{1)} Generates an initial image, \textbf{2)} Uses a VLM Checker to provide CoT reasoning and identify unsatisfied spatial relationships, \textbf{3)} Employs a VLM Actor to think and propose targeted edits, and \textbf{4)} Applies these edits via an Image Editor. This process repeats until all constraints are satisfied or a maximum iteration limit is reached.
%RL-RIG consists of four parts: a VLM Actor, a VLM Checker, an Image Generator, and an Image Editor. The VLM Actor and the Image Editor contain trainable parameters, while the VLM Checker remains frozen. Here, these components do not refer to any specific models, so any VLM, Image Generator and editor is applicable in practice.% 图：三个框架
\subsection{The generation process}

The generation process of RL-RIG begins with an initial image $I = \mathcal{G}(Q)$ produced by a text-to-image Generator $\mathcal{G}$. The VLM Checker, based on a reasoning-finetuned VLM model, evaluates whether the generated image satisfies specified spatial relationships, item by item:
\begin{equation}
    \Gamma_{Passed} = \begin{cases}
        True,  &\text{if } \mathcal{C}(I,Q)=1\\
        False,&\text{otherwise}
    \end{cases}
\end{equation}

If the image fails this evaluation, the VLM Actor, trained from the VLM Checker, will: \textbf{1)} Evaluate the image and find the unsatisfied requirements; \textbf{2)} Analyse the failure reason; \textbf{3)} After CoT thinking, generate an editing prompt:
\begin{equation}
Q^\prime= \mathcal{VLM}_{actor,\phi}(I,Q).
\end{equation}

The Image Editor uses this prompt to modify the image:
\begin{equation}
    I^\prime=  \mathcal{E}_\rho (I, Q^\prime).
\end{equation}
This process iterates through multiple cycles. In each cycle, the Checker re-evaluates the edited image and the system applies further edits as necessary. The process continues until either the image satisfies all spatial criteria or the system reaches a predefined maximum number of editing steps.

The CoT reasoning process in both the VLM Checker and the VLM Actor accurately evaluates unsatisfied requirements and fixes them by image editing. The explicit mechanism enables each part to focus on respective goal and collaboratively perform as a clear pipeline to resolve unsatisfied prompts.

\subsection{The training process}
We observe that relying solely on pretrained modules might be insufficient for generating images that fulfill complex spatial requirements for some cases. This limitation arises from two main issues: \textbf{1)} The VLM Actor lacks knowledge of how to formulate effective edit prompts. \textbf{2)} The Image Editor often struggles to modify images in ways that satisfy all specified relationships, regardless of the prompt provided.  Therefore, \textbf{apart from the framework proposed with appreciable performance gain, we would like to post-train both components to further improve their performance.} Our expected outcome is: \textbf{1)} The VLM Actor should learn a good intuition of writing effective edit prompts; \textbf{2)} Given a well-articulated edit prompt, the Image Editor should know how to perform specific editing operation accordingly. Achieving these goals requires the model to \textbf{compare alternative trajectory branches, assign greater probability to high-advantage ones, and thus refine its ``intuition'' for 1) and 2)}. 

Intuitively, we exploit Reflection-GRPO as a trajectory‑pruning mechanism to discourage unnecessary trajectories while preserving the ones that actually satisfy the given prompts. Group-wise relative advantages upweight edit prompts that raise scene‑graph IoU upon re‑sampling and downweight those that do not.

%To align with the definitions in Reinforcement Learning for GRPO,  
We can formulate the generation framework as a Markov Decision Process (MDP) defined by the tuple \((\mathcal{S}, \mathcal{A}, P, R)\), where the state \(s_t\in\mathcal{S}\) encodes the current image $I$ and target scene graph, the action \(a_t\in\mathcal{A}\) is either an initial generation prompt or an edit prompt, \(P(s_{t+1}\mid s_t,a_t)\) is the (stochastic) image update via the generator or editor, \(R(s_t,a_t)=r_t\) is the spatial‐fidelity reward from the VLM Checker \cite{Sutton2018}.

As shown in \autoref{train}, we propose Reflection-GRPO to post-train RL-RIG. It contains two phases: \textbf{Training the VLM Actor only, and training the Image Editor only}. In both phases, the reward signal is provided by a fixed VLM Checker. % To optimize the components, we employ GRPO which estimates the relative advantage of each candidate response given by the VLM Checker within a sampled group.
\vspace{-1em}
\paragraph{Phase 1} The first phase aims to address issue 1), training the VLM Actor for generating good edit prompts. First, we filter the images in the training set that fail the evaluation. Next, we let a group of $|G|$ VLM Actors generate the edit prompts separately using different random seeds:% \vspace{0.5em}
\begin{equation}
    a_{t}^i=\mathcal{VLM}_{actor,\phi,\xi^i}(s_t,a_0), \quad i\in [G]
\end{equation}
where $i$ is the id of VLM Actor in the group, $a_t^i$ means the edit prompt $a_t$ generated by the $i$-th actor in the group (same meaning for the rest superscript), and $\xi^i$ is the random seed for each actor.

Then, we perform image editing based on each edit prompt, separately:% \vspace{0.5em}
\begin{equation}
    P(s_{t+1}^i|s_t,a_{t}^i)=\mathbb{E}_{\epsilon}[\mathcal{E}_{\epsilon} (s_t, a_{t}^i)=s_{t+1}^i],\quad  i\in [G]
\end{equation}
where $\epsilon$ is the random seed.

After that, we use VLM Checkers to evaluate the image one by one and get the reward score for the RL policy:% \vspace{0.5em}
\begin{equation}
    R(s_t^i,a_t^i):=r_{t}^i = \mathcal{C}(s_{t+1}^i,a_0), \quad i\in [G]
\end{equation}
We want to maximize the expectation of the reward $r_t^i$. In this phase, only $\phi$ is trainable, so the goal is $\max ~\mathbb{E}_{i,\phi}[r_t^i]$. Following the approach of GRPO \cite{guo2025deepseek}, we calculate the advantage and update the trainable parameters in VLM Actor:
\begin{equation}
\phi \;\leftarrow\; \phi \;+\; \alpha \,\nabla_{\phi}\,\mathcal{J}^{\mathrm{GRPO}}(\phi)
\end{equation}% where
\begin{equation}
\small
\begin{aligned}
\mathcal{J}^{\mathrm{GRPO}}(\phi)
&=\frac{1}{G}\sum_{i=1}^{G}
\min\bigl(\rho_{t\_i}\,\widehat{A}_{t\_i},\;\mathrm{clip}(\rho_{t\_i},1-\epsilon,1+\epsilon)\,\widehat{A}_{t\_i}\bigr)
\\&\;-\;\beta\,D_{\mathrm{KL}}\bigl[\pi_{\phi}\!\mid\!\pi_{\mathrm{ref}}\bigr].
\end{aligned}
\end{equation}
Here, $\mu_r = \frac{1}{G}\sum_{j=1}^G r_{t}^j$, $ \sigma_r = \sqrt{\frac{1}{G}\sum_{j=1}^G\bigl(r_{t}^j - \mu_r\bigr)^2}$, $\widehat{A}_{t}^i = \frac{r_{t}^i - \mu_r}{\sigma_r}$, $\rho_i = \frac{\pi_{\phi}(a_{t}^i\mid s_{t}^i)}{\pi_{\mathrm{ref}}(a_{t}^i\mid s_{t}^i)}$, and $\pi$ is $\mathcal{VLM}$ 's probability of taking such action.

\paragraph{Phase 2} In the second phase, we aim to address issue (2) by improving the performance of Image Editor after the VLM Actor performs relatively well. Therefore, we freeze the VLM Actor and train the  Image Editor. After filtering the image, we use one trained VLM Actor to generate the edit prompt, and sample a group of edited trajectories by varying
diffusion noise seeds  of image editors. Then, we use VLM Checkers to evaluate each edited image in group one by one:
% \vspace{0.5em}
\begin{equation}
    a_{t}=\mathcal{VLM}_{actor,\phi,\xi}(s_t,a_0),
\end{equation}
\begin{equation}
s_{t+1}^i = \mathcal{E}_{\rho}(s_t, a_t; \epsilon^i),\quad i\in[G]
\end{equation}
\begin{equation}
    P(s_{t+1}^i|s_t,a_{t}^i)=\mathbb{E}_{\epsilon}[\mathcal{E}_{\rho, \epsilon} (s_t, a_{t}^i)=s_{t+1}^i],\quad  i\in [G]
\end{equation}
where $\xi$ and $\epsilon$ are random seeds, and $\rho$ is the trainable parameter for the Image Editor. Checker rewards are computed as $r_t^i=\mathcal{C}(s_{t+1}^i,a_0)$ and normalized into $\hat A_i$. The training goal of this phase is $\max ~\mathbb{E}_{i,\rho}[r_t^i]$.  
For latent transition step $k$,
 we compute
\begin{equation}
r_{i,k}^{\text{ratio}}=
\exp \bigl(\log p_{\rho}(x_{k-1}^i\!\mid x_k^i,\cdot)-\log p_{\rho_{\text{old}}}(x_{k-1}^i\!\mid x_k^i,\cdot)\bigr),
\end{equation}
and optimize
\begin{equation}
\begin{aligned}
\mathcal{J}^{\text{EDIT}}=&
\frac{1}{GK}\sum_{i=1}^{G}\sum_{k=1}^{K}
\min\!\left(r_{i,k}^{\text{ratio}}\hat A_i,\,
\mathrm{clip}(r_{i,k}^{\text{ratio}},1-\epsilon,1+\epsilon)\hat A_i\right)\\
-&\beta\,D_{\mathrm{KL}}\bigl[\pi_{\phi}\!\mid\!\pi_{\mathrm{ref}}\bigr].
\end{aligned}
\end{equation}
% \begin{equation}
%     r_{t}^i = \mathcal{C}(s_{t+1}^i,a_0),\quad i\in [G]
% \end{equation}

In practice, we update only LoRA parameters in attention projections
$\{\texttt{to\_q},\texttt{to\_k},\texttt{to\_v},\texttt{to\_out.0}\}$,
while keeping base DiT weights frozen. \textbf{We offer more details on training in \autoref{train_details}.}

%% file: sec/4_experiments.tex
\section{Experiments}
\newcommand{\pos}[1]{\cellcolor{green!12}#1}
\newcommand{\negv}[1]{\cellcolor{red!12}#1}
\subsection{Experimental Settings}

\paragraph{Base models} In practice, we use Qwen-3-VL-Thinking \cite{qwen3technicalreport} as the base VLM model for a better reasoning ability. The Image Generator employs Flux \cite{flux2024}, a state-of-the-art open-source image generation model, and the Image Editor employs Qwen-Image-Edit \cite{wu2025qwenimagetechnicalreport}, enabling efficient editing of real images based on given text descriptions without requiring a mask area. The GRPO training framework is modified from VLM-R1 \cite{shen2025vlm, openr1}. In the future, all these components can be substituted by a stronger model if they are available for the same task. More details can be found in \autoref{sec:experiment_detail}.

\begin{table*}[t]
\setlength{\tabcolsep}{4pt}
    \centering
    \caption{Comparison of different models on GenEval dataset.  The highest score for each metric is \textbf{bolded}.}
    \resizebox{0.8\textwidth}{!}{%
    \begin{tabular}{c|c|c|c|c|c|c|c}
    \toprule
       \textbf{Name} & \textbf{Overall} & \textbf{1 object} & \textbf{2 objects} 	& \textbf{Counting} & \textbf{Colors}  & \textbf{Position} & \textbf{Color attribution}  \\
    \midrule
    CLIP retrieval &	0.35 	& 0.89 &	0.22 	& 0.37 	& 0.62  &	0.03 &	0.00 \\
minDALL-E 	& 0.23  &	0.73 &	0.11 	& 0.12 &	0.37 &	0.02 	& 0.01 \\
SD v1.5 	& 0.43 & 	0.97 & 	0.38 	& 0.35 	& 0.76 	& 0.04 	& 0.06 \\
SD v2.1 	& 0.50  &	0.98 & 	0.51 	& 0.44 	& 0.85 	& 0.07 	& 0.17 \\
      SD-XL & 0.55 & 0.98 & 0.74 & 0.39 & 0.85 & 0.15 & 0.23 \\
IF-XL  & 0.61 & 0.97 & 0.74 & 0.66 & 0.81 & 0.13 & 0.35 \\
\rowcolor{cyan!10} \textbf{RL-RIG} & \textbf{0.95} & \textbf{0.99} & \textbf{0.96} & \textbf{0.94} & \textbf{0.98} & \textbf{0.85} & \textbf{0.97}\\
 \bottomrule
    \end{tabular}
    }
    \label{tab:geneval}
\end{table*}

\begin{table*}[t]
\centering
\caption{Comparison of different models on LAION-SG dataset. The variance is obtained by running the judger 10 times. The highest score for each metric is \textbf{bolded}, and the second highest score is \underline{underlined}.}

\label{tab:sg-evaluation}
\setlength{\tabcolsep}{1pt}
\resizebox{\textwidth}{!}{%
\begin{tabular}{lccccc}
\toprule
\textbf{Method} & \textbf{SG-IoU} & \textbf{Ent-IoU} & \textbf{Rel-IoU} & \textbf{Qwen-Judge} & \textbf{GPT-Judge} \\
\midrule
SD3.5 Large & 0.3135 ± 0.0071 & 0.8788 ± 0.0010 & 0.7650 ± 0.0056 & 0.7785 ± 0.0021 & 0.7747 ± 0.0046 \\ 
Flux & 0.3461 ± 0.0078 & 0.9079 ± 0.0011 & 0.7695 ± 0.0074 & \underline{0.8012 ± 0.0018} & \underline{0.8100 ± 0.0045} \\ 
\rowcolor{gray!20} Flux + RL & 0.3445 ± 0.0057 & 0.9214 ± 0.0026 & 0.7937 ± 0.0047 & 0.6647 ± 0.0085 & 0.6484 ± 0.0034 \\ 
LAION-SG & 0.2693 ± 0.0060 & 0.7789 ± 0.0023 & 0.6811 ± 0.0069 & 0.6360 ± 0.0043 & 0.6331 ± 0.0032 \\ 
\rowcolor{cyan!10} RL-RIG (w/o RL) & \textbf{0.3702 ± 0.0097} & \underline{0.9319 ± 0.0033} & \underline{0.8006 ± 0.0053} & 0.7931 ± 0.0048 & 0.7648 ± 0.0048 \\ 
\rowcolor{cyan!10} \textbf{RL-RIG} (w RL) & \underline{0.3644 ± 0.0077} & \textbf{0.9365 ± 0.0031} & \textbf{0.8015 ± 0.0065} & \textbf{0.8269 ± 0.0053} & \textbf{0.8401 ± 0.0063} \\ 
\midrule
\rowcolor{gray!20} RL-RIG (w/o reflection) @10  & 0.3468 & 0.9156 & 0.7729 & 0.8392 & 0.7273 \\
Ground Truth & 0.3864 & 0.9304 & 0.8129  & 0.7240 & 0.7536\\
\bottomrule
\end{tabular}%
}
\end{table*}

\paragraph{Dataset} To ensure a standard and comparable evaluation, we first assess RL-RIG on the GenEval benchmark \cite{ghosh2023geneval}, where it achieves near-perfect performance. This suggests that datasets like this are too easy for our current model, because they do not contain such complex spatial relationships in the prompt, and evaluate outputs with at most two or three simple objects. Consequently, we turn to LAION-SG \cite{li2024laion} for training and testing, which is a remarkably challenging dataset specifically curated for its \textbf{highly intricate spatial relationships among multiple objects}.

The data split is set by default. For convenience, we select top 2000 images for training by the aesthetic score.  We use the same approach as described in LAION-SG to utilize GPT-4o for extracting the scene graph and evaluating the IoU. 
\paragraph{Evaluation metrics} We observe that ground truth images do not align perfectly with the provided spatial descriptions, and thus applying traditional metrics like FID and IS would lead to biased results. Instead, we use metrics like Scene Graph IoU \cite{zhang2024multiview} and VLM-as-a-judge \cite{pan2025prometheus} that can evaluate the images' loyalty to the extracted relationships. Suppose $Q$ is the Scene Graph extracted from ground truth (i.e., the text requirements), and $\hat{Q}$ is the Scene Graph extracted from the generated image by GPT-4o \cite{openai2024gpt4o}. Then, Scene Graph IoU is defined as the IoU between them:
\begin{equation}
    IoU_{SG}=\frac{Q \cap \hat{Q}}{Q \cup \hat{Q}}
\end{equation}
Suppose $O=\{o_i \in Q\}$ is all the objects that appear in $Q$, and $\hat{O}=\{\hat{o_i} \in \hat{Q}\}$ is all the objects that appear in $\hat{Q}$. Similarly, $Rel=\{r_i|(o_{x_i},r_i,o_{y_i}) \in Q\}$ and $\hat{Rel}=\{\hat{r_i}|(\hat{o_{x_i}},\hat{r_i},\hat{o_{y_i}}) \in \hat{Q}\}$ is all the relationships in generated image and ground truth image, separately. Then, the entity IoU and relationship IoU are defined as:
\begin{equation}
    IoU_{Ent}=\frac{O \cap \hat{O}}{O \cup \hat{O}},\quad IoU_{Rel}=\frac{Rel \cap \hat{Rel}}{Rel \cup \hat{Rel}}
\end{equation}
\textbf{Unlike traditional metrics that measure image similarity, our IoU-based evaluation specifically targets spatial relationship fidelity.} This approach enables us to differentiate between images that merely appear visually similar to ground truth and those that accurately represent the specified spatial relationships. In contrast, traditional pixel-level metrics like IS and FID fail to capture the critical spatial configurations that define success in our problem setting.

The score of VLM-judge is defined in the same way as \autoref{checker}. Specifically, LLM would evaluate the number of relationships from the prompt that is satisfied in the picture, which emphasizes faithfulness to the prompt instead of the “ground truth” pictures most models use. \textbf{Instinctively, we believe there are no ground truth pictures in image generation. Any picture that satisfies the requirements in the prompt should be well accepted.} Here, we employ Qwen3-VL-Thinking and GPT-4o as two VLM judges for meta-evaluation, separately.

\vspace{-1em}
\subsection{Results}
We choose SD3.5 Large, Flux and LAION-SG as baseline models. The former two models are state-of-the-art open-source image generation models; the latter one is the model from the original literature of the LAION-SG, which is specially post-trained on the LAION-SG dataset. We compare the performance of RL-RIG before and after post-training to better understand its influence.  All comparisons are conducted using the same original prompts from the LAION-SG dataset without any modification.

The result of our experiment is shown in \autoref{tab:geneval} and \autoref{tab:sg-evaluation}. RL-RIG consistently outperforms all baselines across spatial reasoning metrics. Notably, our method shows the largest improvements on the most challenging spatial relationships (scene graph), validating our iterative refinement approach. The RL+Flux baseline results in a better IoU but also a significant drop in ACC. We reckon that CLIP limits FLUX's ability to understand complex spatial requirements in the text.

\paragraph{Efficiency} The average edit rounds is 2.76. When pipelined on 4*A100s, the average time is  26s/image, 1.90x slower than Flux but 5.27x faster than `Pick 1 from 10'. \textbf{It is acceptable for real-world applications.} When \texttt{max\_edit\_round=5}, the termination rate is 48.4\%, which means 51.6\% of the images have acquired a perfect 1.0 score from the VLM Judger. (For the rest, the round that achieves the highest score will be picked.)

\begin{figure}[htbp]
	\centering
	\vspace{-1em}\includegraphics[width=\linewidth]{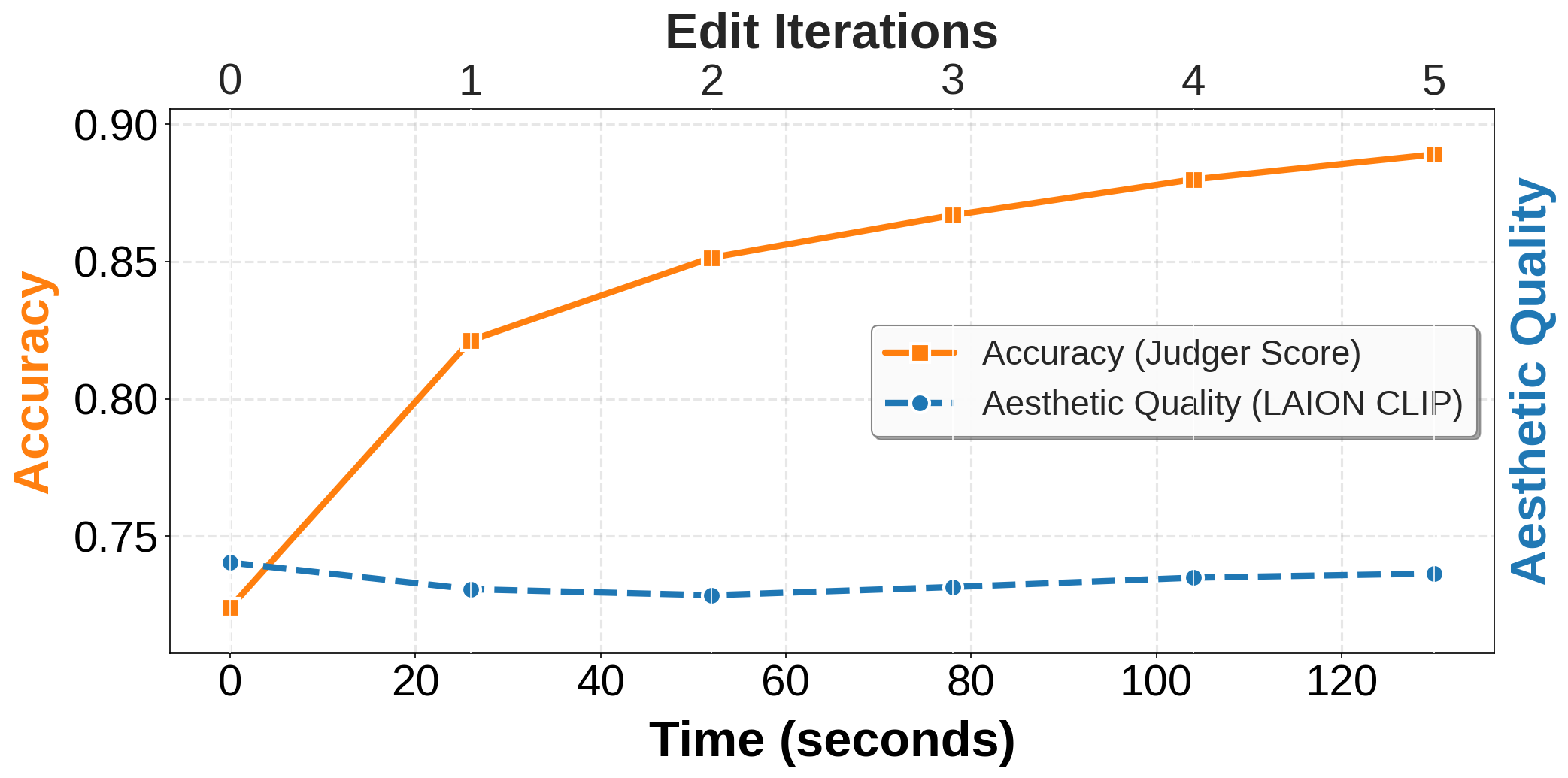}
	\caption{The relationship between edit iterations, time, accuracy, and aesthetic quality.}
 \vspace{-2em}
	\label{img_dual_metrics}
\end{figure}

\paragraph{Judge Reliability and Sensitivity} \cref{tab:sg-evaluation} show the multi-run std is relatively low, which indicates the stability of our VLM judge's results. We conduct correlation test: Intra-judge correlation within tests is 0.8003 for GPT and 0.7826 for Qwen. The inter-judge correlation between them is 0.5873. Therefore, same judges tend to provide same scores, and different judges' scores can align well. Human audits have reviewed the judges' response and find that in $\sim 80\%$ cases, judges align well with human audit. In some of the few cases below where judges disagree, they are ambiguous to human audit too:
\vspace{-1em}
\newcommand{\uncertain}[1]{\colorbox{yellow!25}{#1}}
\begin{figure}[H]
\centering
\resizebox{\linewidth}{!}{%
\large
% \fontsize{40pt}{14pt}\selectfont
\setlength{\tabcolsep}{0pt}
\begin{tabular}{m{0.25\textwidth}|m{0.25\textwidth}|m{0.25\textwidth}|m{0.25\textwidth}}
\centering
\includegraphics[width=\linewidth]{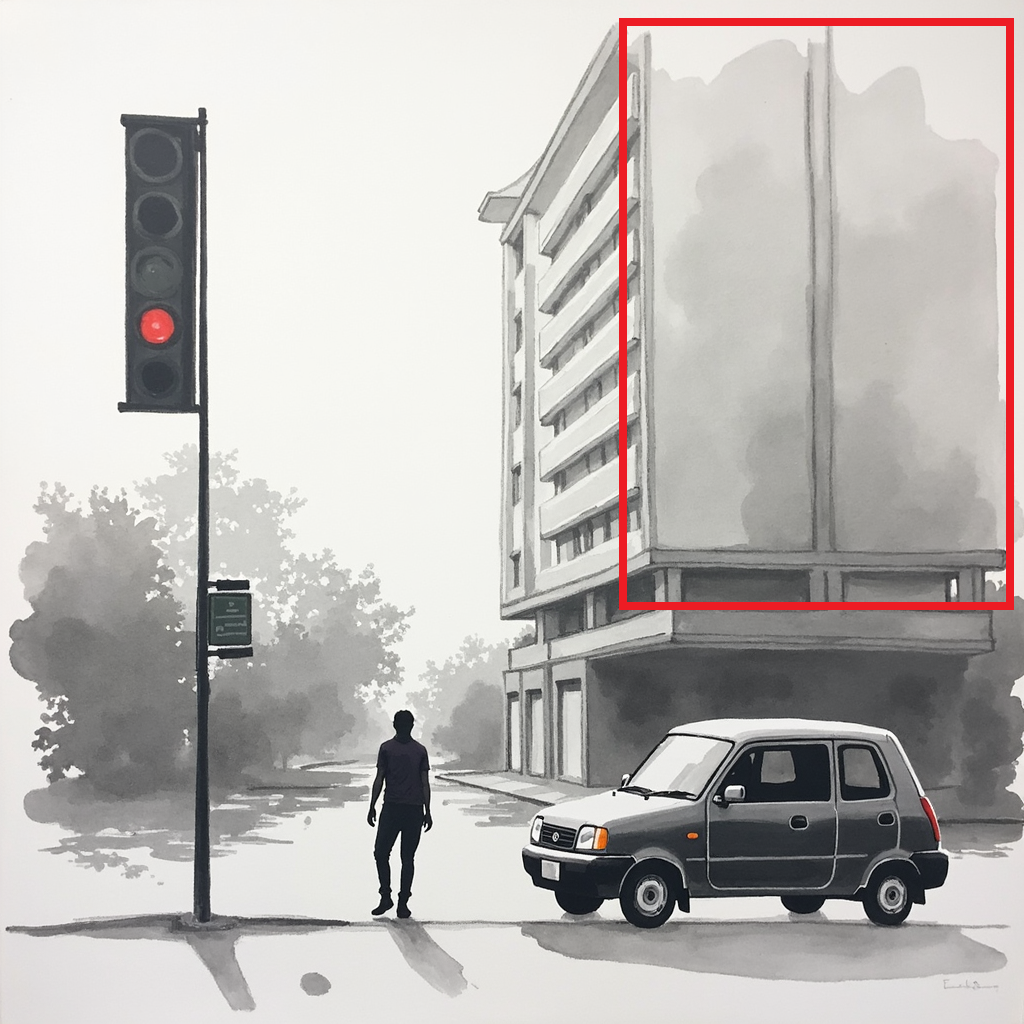} & \includegraphics[width=\linewidth]{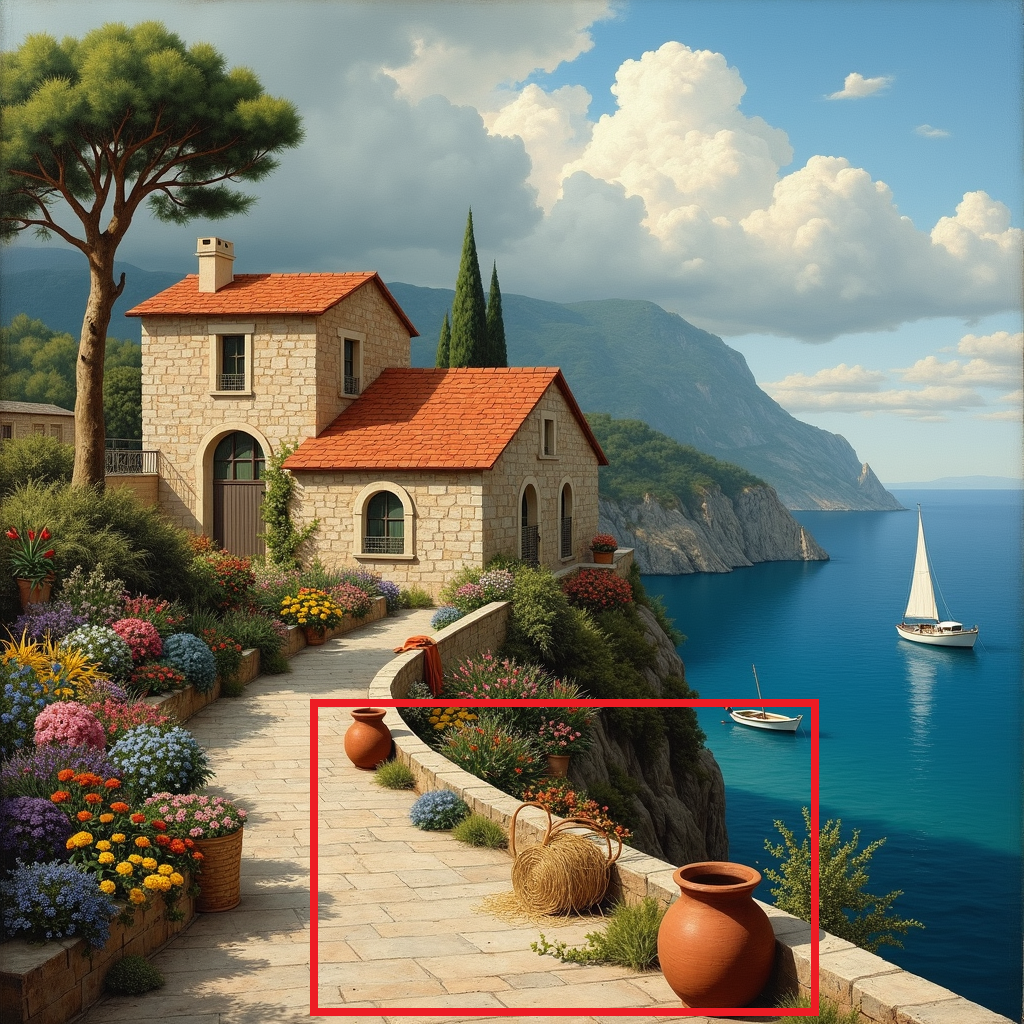} & \includegraphics[width=\linewidth] {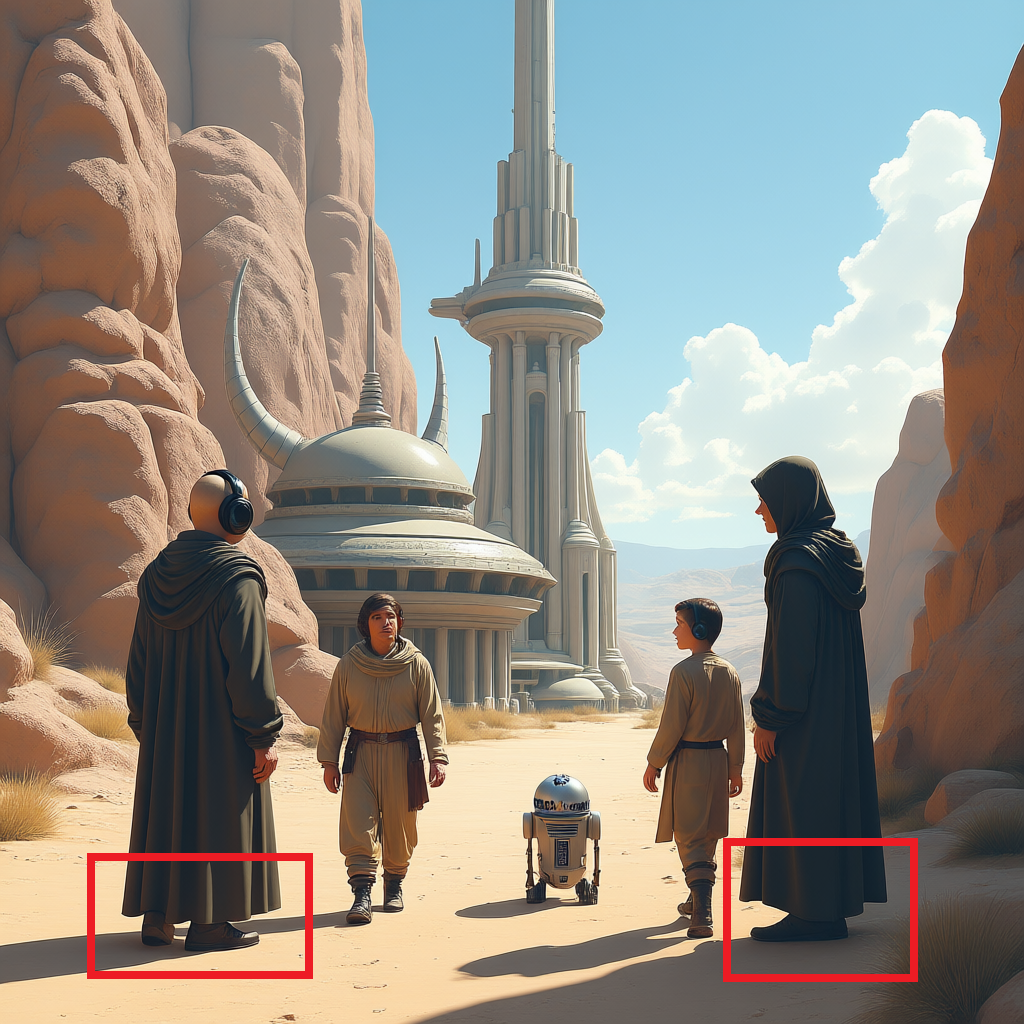} & \includegraphics[width=\linewidth] {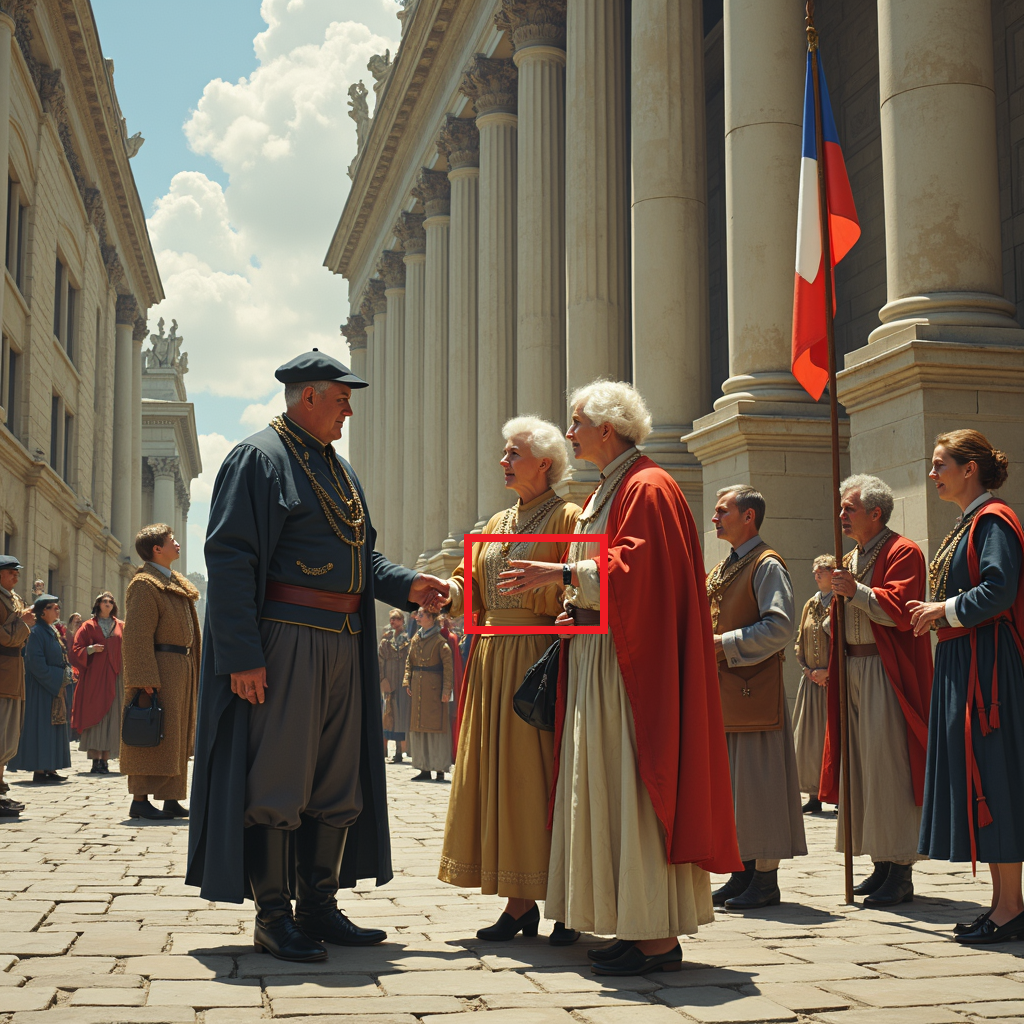}
\\
\makecell{Large \emph{\uncertain{white(?)}} building;} &   \makecell{Clay vase \emph{\uncertain{placed on(?)}}\\ stone red-roofed house;} &  \makecell{Tall cloaked person \\\emph{\uncertain{walking(?) toward}}\\ spiky tall structure; } &  \makecell{Female middle-aged \\ person \emph{\uncertain{pointing at}} \\ male young person;}
\end{tabular}
}
\vspace{-0.5em}
\caption{Ambiguous generation examples where judges disagree.}
\end{figure}
\paragraph{Accuracy–Efficiency Trade-offs} We study the relationship between maximum edit steps, time, accuracy, and aesthetic quality (measured by LAION CLIP) in \autoref{img_dual_metrics}.  The aesthetic loss comes from the image editor and is more or less inevitable, while we believe satisfying the input prompt is prior to pursuing aesthetic qualities. Nevertheless, \autoref{img_dual_metrics} shows RL-RIG still achieves strong aesthetic performance (measured by LAION aesthetics) when the edit step increases. The aesthetic loss is not universal.

\subsection{Case Study}
\label{case}

We provide three groups of comparison of the image generated by these models in \autoref{fig:comparison}. It can be observed that RL-RIG performs a stronger ability of following the extremely complex spatial descriptions in these examples, accurately capturing relationships that other models struggle with.

\begin{figure}[htbp]
\vspace{-1em}
  \centering
    \scriptsize
   \begin{tcolorbox}[colback=white, colframe=black, boxrule=0.8pt, sharp corners, boxsep=0pt]
\textbf{Prompt:}{ \textit{ \textbf{\textcolor{blue}{driving person riding two-wheeled vehicle}}; tall old building adjacent to tall old building; \textbf{\textcolor{blue}{black car vehicle following two-wheeled vehicle}}; yellow taxi vehicle parkedside yellow taxi vehicle; tall old building adjacent to tall bright building; small sign attached to tall old building; A new day, Mumbai by Ramesh Jhawar, Impressionism Painting, \textbf{\textcolor{blue}{Watercolor}} on Paper, Brown color}} %\quad     \textbf{ID:} 523378
  \end{tcolorbox}
\begin{tabular}{ 
    >{\centering\arraybackslash}m{1.3cm}  % 左列：水平+垂直居中
    | p{6.7cm}                            % 右列：默认顶对齐、左对齐
}
\toprule
\textbf{Agent} & \multicolumn{1}{c}{\textbf{Response}}  \\
\midrule
\textbf{Generator} & Generated with the original prompt\\
\midrule
\makecell{\textbf{Checker}} & <think> Let's analyze each description one by one: {\begin{enumerate}[left=0pt,nosep] 
    \item \textit{``black car vehicle following two-wheeled vehicle''}: There is no two-wheeled vehicle in the image, so this description is not satisfied.
    \item \textit{``yellow taxi vehicle parkedside yellow taxi vehicle''}: There are two yellow taxis parked next to each other, so this description is satisfied.
    \item \textit{``small sign attached to tall old building''}: There is a small sign visible on one of the tall buildings, so this description is satisfied.
    \item \textit{``tall old building adjacent to tall old building''}: There are two tall buildings next to each other, so this description is satisfied.
    \item \textit{``driving person riding two-wheeled vehicle''}: There are no two-wheeled vehicles in the image, so this description is not satisfied.
    \item \textit{``tall old building adjacent to tall bright building''}: There is a tall old building next to a taller bright building, so this description is satisfied.
\end{enumerate}}  </think><answer> \boxed{4}</answer>\\
\midrule
\makecell{\textbf{Actor}} & <think>...</think><answer>{\textit{yellow taxi vehicle parked, small sign attached to tall old building, tall old building adjacent to tall old building, tall old building adjacent to tall bright building, black car vehicle following two-wheeled vehicle, watercolor}}</answer>\\
\midrule
\textbf{Editor} & Performed  editing with the given prompt\\
\midrule
\textbf{Checker} & <think>...</think><answer> \boxed{6}</answer>\\
\bottomrule
\end{tabular}
\vspace{0.5em}

  \begin{minipage}{0.13\textwidth}
        \centering
        \includegraphics[width=\textwidth]{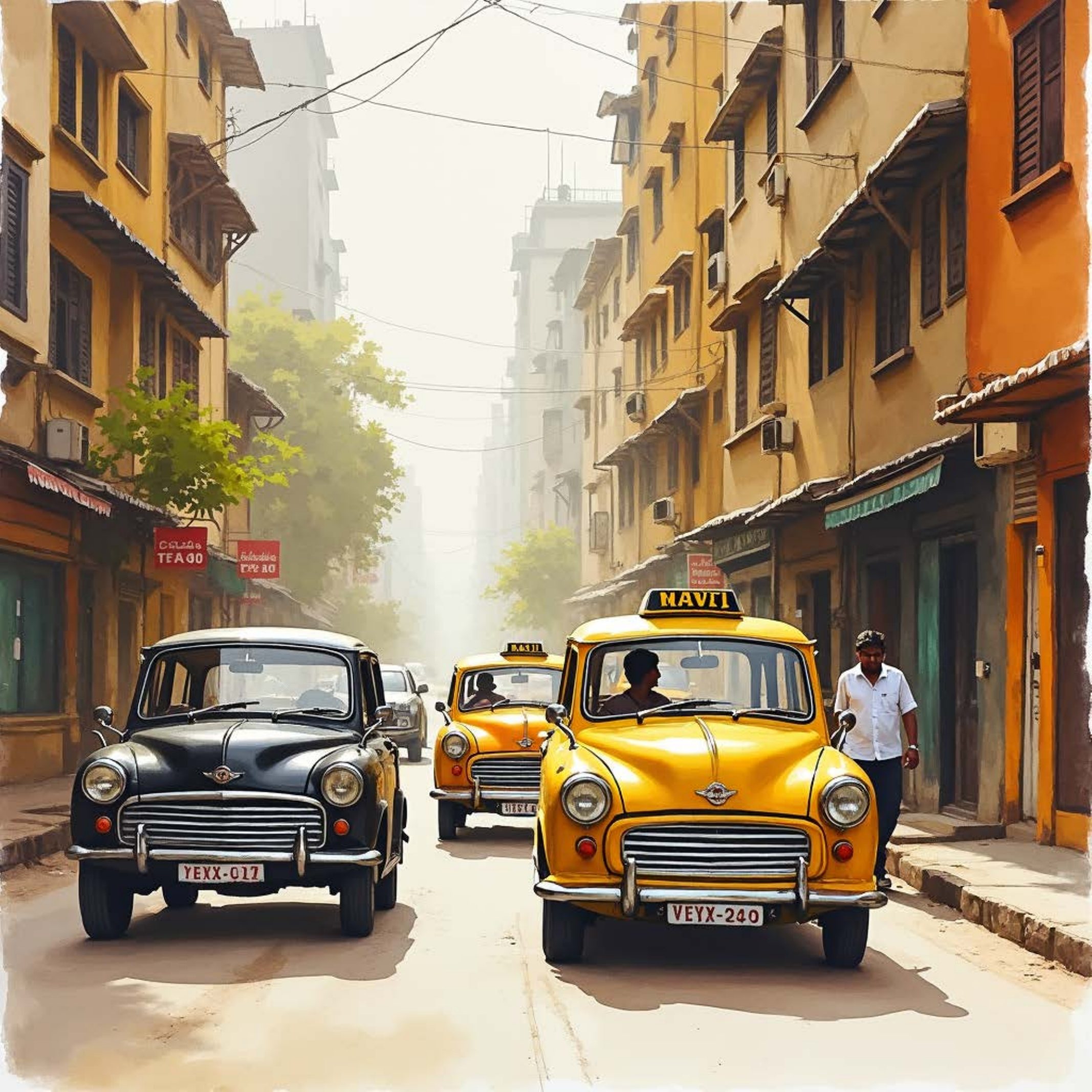}
    \end{minipage} 
  \begin{minipage}{0.05\textwidth}
        \centering
\includegraphics[width=\textwidth]{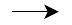}
    \end{minipage} 
    \begin{minipage}{0.13\textwidth}
        \centering
        \includegraphics[width=\textwidth]{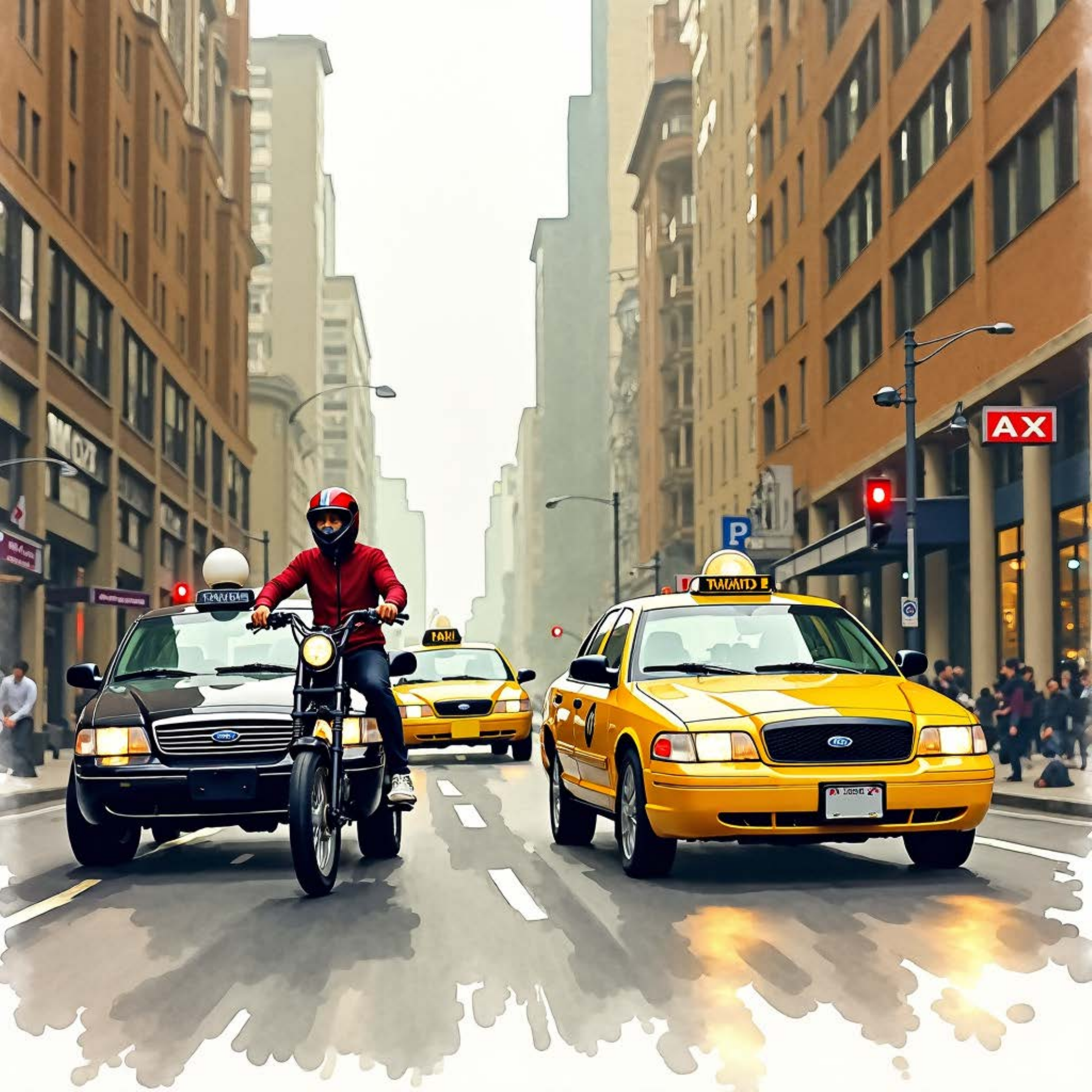}
    \end{minipage}
  \caption{A succeeded trial of image generation by RL-RIG, with the input prompt (id=523378) and the reasoning process. After reflection, the actor successfully guides the Image Editor to add a two-wheeled vehicle in front of the black car.}
      \label{case_success}
    \vspace{-2em}
  \end{figure}

To better illustrate our framework's reasoning process, we present a detailed examination of the intermediate steps in two RL-RIG generation cases using the same prompt, which is shown in \autoref{case_success}. 

In the successful case in \autoref{case_success}, we can observe the complete Generate-Reflect-Edit cycle in action. The initial image generated by the Image Generator captures several spatial relationships correctly, but crucially misses two key requirements: the two-wheeled vehicle in front of the black car, and a person riding it. Next, the VLM Checker meticulously analyzes this initial output through a structured reasoning process, and identifies the missing relationships. The VLM Actor then formulates a targeted edit prompt that preserves correct elements while specifically requesting the missing components. Finally, the Image Editor successfully incorporates these elements, resulting in an image with all spatial requirements.

In contrast, under identical conditions but with a different random seed, \autoref{case_bad} in the Appendix illustrates a failed attempt where the editor either completely overwrites the original image, resulting in a fundamentally different scene, or makes no changes at all. This comparison highlights that our RL-RIG design is capable of reasoning and correcting unsatisfactory outputs from the base model, whereas its performance remains partially constrained by the limitations of the base model itself.

\textbf{We provide more ablation studies in \autoref{sec:ablation}}.

% The result of our experiment is shown in \autoref{tab:sg-evaluation}, where RL-RIG performs better than other baseline models. We observe that, compared with the original base model Flux as well as other baselines, the application of both RL-RIG's framework and post training phases contribute to an increase of performance, which shows the effectiveness of our design. We observe a decrease in Ent-IoU after post-training, which might be attributable to the limited ability of the Image Editor (\textbf{see success and failure case study in \autoref{case}}).
\begin{figure*}[htbp]
    \centering
    \vspace{-1em}
    \includegraphics[width=\linewidth]{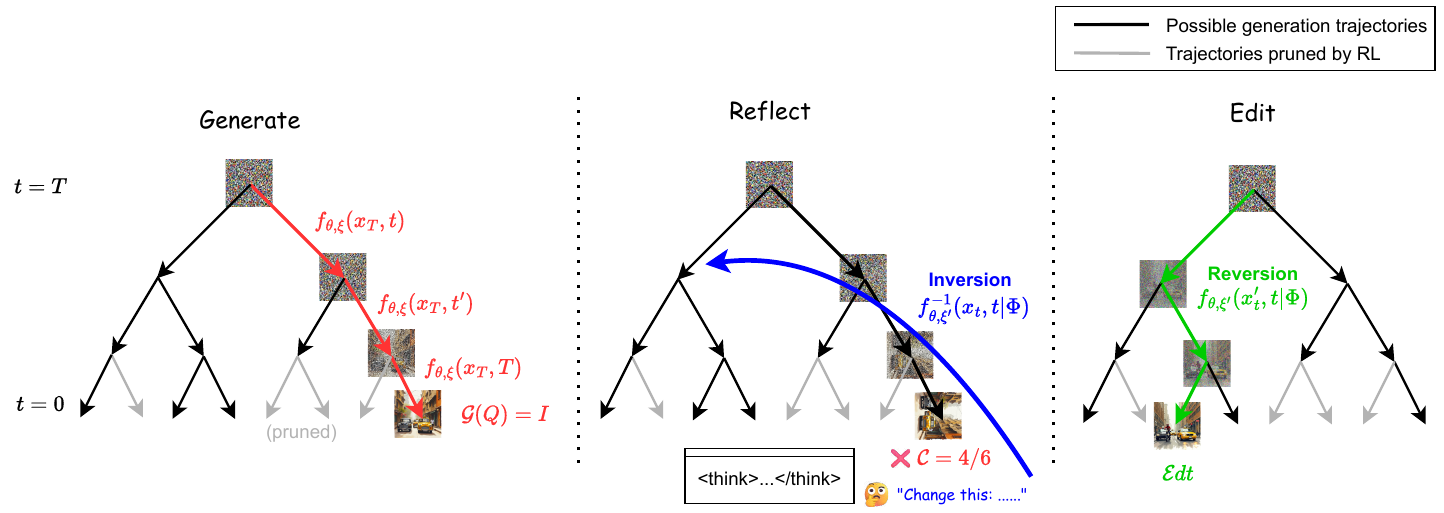}
    \vspace{-1.5em}
    \caption{The Generate-Reflect-Edit framework, explained in a trajectory view. In each generation process, one of the possible trajectories is selected according to the random seeds. The VLM Checker will then reflect and check whether all parts of the prompt are satisfied. If not, the VLM Actor will provide an edit prompt and pass it to editor. Given the edit prompt, the Image Editor will perform inversion and reversion to explore a new possible trajectory. Dashed branches denote low‑advantage trajectories pruned by GRPO. The `Inversion' is actually performed in Edit stage. }
    \label{tree}
\end{figure*}

\subsection{Mechanism: Intrinsic Reflection as Trajectory‑Level Reward}
Where does the performance gain come from? We add another ablation study, `pick 1 from N', by disabling the multiple cycle and post training of RL-RIG, simply generating 10 images and picking the best one by VLM Checker. Although this approach is extremely computationally expensive and impractical in real-world application, it simulates the sampling of 10 possible  trajectories. We find that this `pass@10' result is comparable to that of RL-RIG's pass@1. Following the approach of \cite{yue2025doesreinforcementlearningreally}, 
we believe that our approach teaches the model to intuitively choose a better trajectory by its intrinsic reflection ability, and provide our explanation as follows.

\emph{Intrinsic Reflection} refers to the phenomenon where LLMs can learn to reason by only recursively optimizing self-generated feedback \cite{rafailov2024directpreferenceoptimizationlanguage,wang2025implicitrewardbridgeunified}. This works because \cite{li2025generalistrewardmodelsinside} suggests that reward modeling stage can be replaced by eliciting the knowledge already captured during pre-training. In practice, RLIF \cite{zhao2025learningreasonexternalrewards} and RLPR \cite{yu2025rlprextrapolatingrlvrgeneral} replace external rewards with the model’s own confidence or token-likelihood signals while still improving LLM's performance. This suggests that LLMs already embed a \emph{generalist reward model} on thoughts and responses' trajectories.

Our design \textbf{extends this idea to the multi-agent VLM setting}. RL-RIG's \emph{Generate-Reflect-Edit} loop converts internal evaluative signals into implicit rewards purely by VLM Checker's capability of judgment instead of external gold verifiers. The \emph{VLM Checker} computes a scalar \emph{intrinsic reward} $R(s_t^i,a_t^i)$ from the agreement between \textbf{(i)} the prompt-induced scene graph $a_0$ and \textbf{(ii)} the image’s relations $s^i_{t+1}$. In effect, this reward process acts as ``reflection'', supplying a signal to push the policy to increase the probability of edits that raise scene-graph IoU and reduce relation violations.

Inspired by the idea in \cite{rafailov2024rqlanguagemodel}, we can view the model as a latent Q-function over action sequences, where all possible trajectories form a tree, and each diffusion or edit step is actually choosing a branch on it. Based on that, our goal is to couple it with GRPO’s group optimization and optimize the whole system multimodally to choose a better branch.
In RL-RIG,  the reflection step measures relational structure, the intrinsic reward targets it, and GRPO amplifies successful local edits into globally consistent layouts, which eventually  \textbf{sparks the base model's intrinsic ``subconsciousness'' to choose a rarely visited reasoning trajectory that actually performs better for spatial relations.} This internal preference-based training can therefore deliver structurally faithful images rather than merely photorealistic ones.

%% file: sec/5_limitations.tex
\section{Limitations and Future Work}
While our proposed RL-RIG framework demonstrates impressive capabilities, its performance is naturally influenced by the underlying base model, though all the post-training processes can alleviate this influence. Additionally, due to the pioneering nature of our work in addressing complex prompt reasoning, a comprehensive text-to-image dataset for such advanced and difficult spatial relationships is notably rare and remains in development. %The GRPO training approach, though highly effective for our specific scenario, is computationally intensive, which led us to use fewer training and testing samples than conventional approaches.

Our research opens several promising avenues for future work. First, integrating more advanced reasoning mechanisms could further enhance the VLM Actor's ability to generate precise edit prompts. Second, developing specialized Image Editors that maintain better contextual consistency across multiple edits could address the discontinuity issues observed in \autoref{case_bad}. Finally, creating a dedicated dataset for text-to-image tasks with extremely complex  spatial relationships would greatly help this future work.

%% file: sec/6_conclusion.tex
\section{Conclusion}
In this paper, we addressed the ``spatial reasoning dilemma'' in text-to-image generation by proposing RL-RIG, a novel Generate-Reflect-Edit  framework that combines Chain-of-Thought reasoning with image generation to enhance test-time scaling spatial reasoning ability for complex scene generation. We also employ group-reward based reinforcement learning to vision-language models, which frames spatial image generation as trajectory selection. We show that intrinsic reflection yields a better intuition, which is practical to prune low‑fidelity branches and realize scene‑graph‑faithful images.  Experimental results on the LAION-SG dataset demonstrate that RL-RIG outperforms state-of-the-art models like Stable Diffusion 3.5 Large and Flux in terms of spatial coherence, as measured by metrics like Scene Graph IoU  and VLM-based evaluation. RL-RIG is flexible in that all its components can be substituted by the latest state-of-the-art model.

We hope this work highlights the importance of reasoning in complex image generation tasks, and provides an insightful intrinsic reflection  framework to overcome the complex spatial relationships in such scenarios.

%% file: sec/X_suppl.tex
\clearpage
\appendices
\section{More Details of Framework Design}
 The separate design of the VLM Actor and Checker is on purpose.  In practice, we observed that the VLM model is easily distracted when assigned with multiple tasks. Specifically, we tested on the VLM Actor and found that, when required to provide an edit prompt, the VLM Actor (either post-trained or not post-trained) tends to assume that `the image already satisfies all descriptions, and does not need editing`, even if the image actually does not comply with the prompt. However, when the request for editing is dropped, the VLM Actor will move its reasoning focus to checking the satisfied request, which greatly improves the upcoming VLM Checker’s response quality. 

 It is also not appropriate to eliminate the use of the VLM Editor and keep feeding the same prompt, which stems from two main factors. First, inversion-based image editing relies on guidance from the input text prompt. Repeatedly feeding the same prompt leads to identical guidance signals, resulting in little to no improvement over single-pass editing. Moreover, each round of inversion and reversion introduces additional noise, which can degrade image quality. Second, the image editor is highly sensitive to the style of the edit prompt. In practice, directly reusing the original prompt often fails to produce the desired modifications, as it may not align well with the editor’s preferred input style.
\section{More Details of Two-Phase Training}
\label{train_details}
\begin{figure*}[bp]
    \centering
\includegraphics[width=\linewidth]{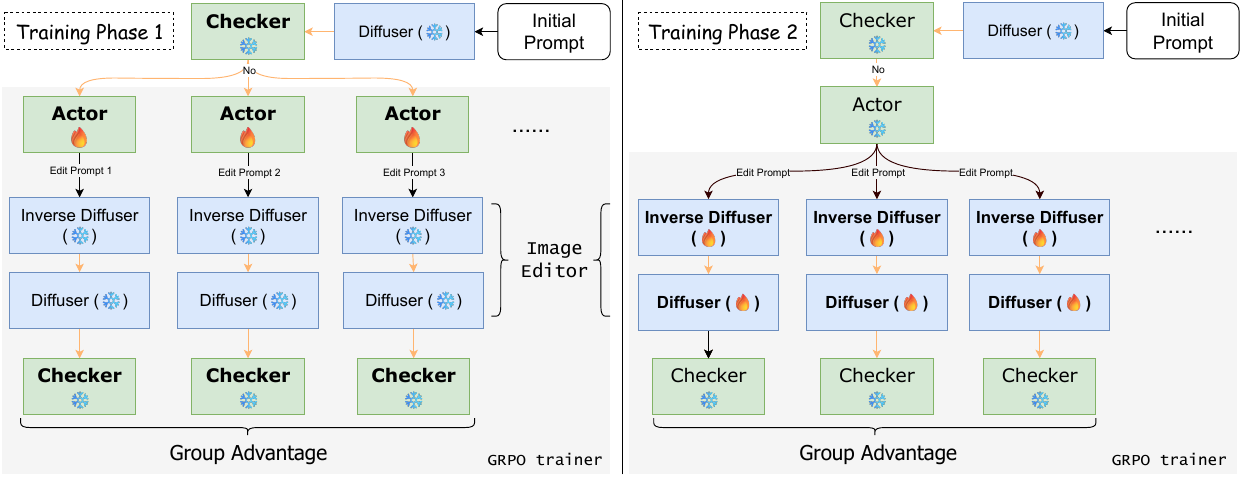}
\vspace{-2em}
    \caption{Illustration of two-phase training. For each phase a batch of responses are sampled, and group advantage is calculated by GRPO.}
    \label{train}
\end{figure*}

Our optimization only includes 2 phases, as is shown in \autoref{train}.

In the first training phase, the VLM Actor generates a batch of edit prompts. Each edit prompt is passed to a frozen Image Editor, and the corresponding image is graded by a VLM Checker. The checker’s response is used as the reward function in GRPO to encourage it to generate an appropriate edit prompt.

In the second training phase, the VLM Actor is frozen, and a fixed edit prompt is passed to the Image Editor to generate a batch of image edits. The corresponding image is graded by a VLM Checker, whose response is used as the reward function in for refining the Image Editor’s performance under the given prompt.

 For the sake of generalization, we keep the VLM Checker untrained. This is because training it would bring about unnecessary bias to its feedback results. Once influenced by the training objective, the reward system would become distorted and overfit to its bias. To further confirm the reliability of the base VLM model, we manually examined 20 cases and found that most of VLM checkers’ outputs are aligned with human senses. Therefore, we conclude that the base VLM model shows a good zero-shot ability with regard to judging how many requirements are satisfied in the image, and that additional post-training on the VLM Checker would be potentially redundant.
 
 The VLM Actor, on the other hand, does require post-training due to its functions in the system. We expect to yield edit prompts from the actor that could bring about images with exact spatial relationships, but we find that a tiny change in edit prompts would result in significantly different results. Notably, even a human-written edit prompt may not yield the desired result in many cases, sometimes performing worse than the post-trained VLM Actor, This shows the necessity of post-training the VLM Actor so as to cooperate well with the features of the Image Editor, its partner in the image generation process, as well as sparking its intrinsic reasoning ability to yield the correct intuition among possible trajectories.

 Although we currently use \emph{split prompts} for better interpretability, the VLM Checker is still a generic reasoning model and can assess the degree of prompt satisfaction through few-shot prompting, even without explicit decomposition. According to \cite{deng2025boostinggeneralizationreasoningvision}, the base VLM model performs well on multiple tasks including image QA, detection, classification and maths, which proves its generalization capabilities. Our training process is not detrimental to the VLM Checker and thus the input is actually not limited to multi-point split prompts, but generalizes to a broader text-to-image task domain.

\section{More Details of Experimental Settings}
\label{sec:experiment_detail}
\paragraph{Environment Setup.} We set up a distributed training on $8$ A100 GPUs. We preprocess all the image generation process before the training begins. For training phase 1, we load $3$ parallel VLM Actors and $1$ VLM reference model, $1$ VLM Checker and $3$ parallel image editors. For training phase 2, we load $1$ VLM Actor, $3$ parallel image editors, $1$ VLM Checker, $1$ Image Editor, and $1$ reference Image Editor, and $1$ trainer. The training time for each phase is about $2$ days. All the parallel Actors and Editors are the same networks, acting as load balancers. 

\paragraph{Hyper-parameters.} We set most the base models' hyper-parameters by default. In the generation phase, each picture can be edited at most 5 times. If it still fails to pass the checker after $5$ edits, the attempt that achieves the best score will be saved. The optimal number of edit prompts is determined empirically: increasing the count from 5 to 6 yields no significant performance gain, while reducing it to 4 results in diminished performance.

\paragraph{Case Study.} We provide three groups of comparison of the image generated by these models in \autoref{fig:comparison}. It can be observed that RL-RIG performs a stronger ability of following the extremely complex spatial descriptions in these examples, accurately capturing relationships that other models struggle with.

To better illustrate our framework's reasoning process, we present a detailed examination of the intermediate steps in two RL-RIG generation cases using the same prompt as described in \autoref{case_success}. 

In the successful case in \autoref{case_success}, we can observe the complete Generate-Reflect-Edit cycle in action. The initial image generated by the Image Generator captures several spatial relationships correctly, but crucially misses two key requirements: the two-wheeled vehicle in front of the black car, and a person riding it. Next, the VLM Checker meticulously analyzes this initial output through a structured reasoning process, and identifies the missing relationships. The VLM Actor then formulates a targeted edit prompt that preserves correct elements while specifically requesting the missing components. Finally, the Image Editor successfully incorporates these elements, resulting in an image with all spatial requirements.

\begin{figure*}[t]
  \centering
  \begin{minipage}{0.2\textwidth}
        \centering
        \includegraphics[width=\textwidth]{img/edit_1_restart0_523378.pdf}
    \end{minipage} 
  \begin{minipage}{0.05\textwidth}
        \centering
\includegraphics[width=\textwidth]{img/arrow.pdf}
    \end{minipage} 
    \begin{minipage}{0.2\textwidth}
        \centering
        \includegraphics[width=\textwidth]{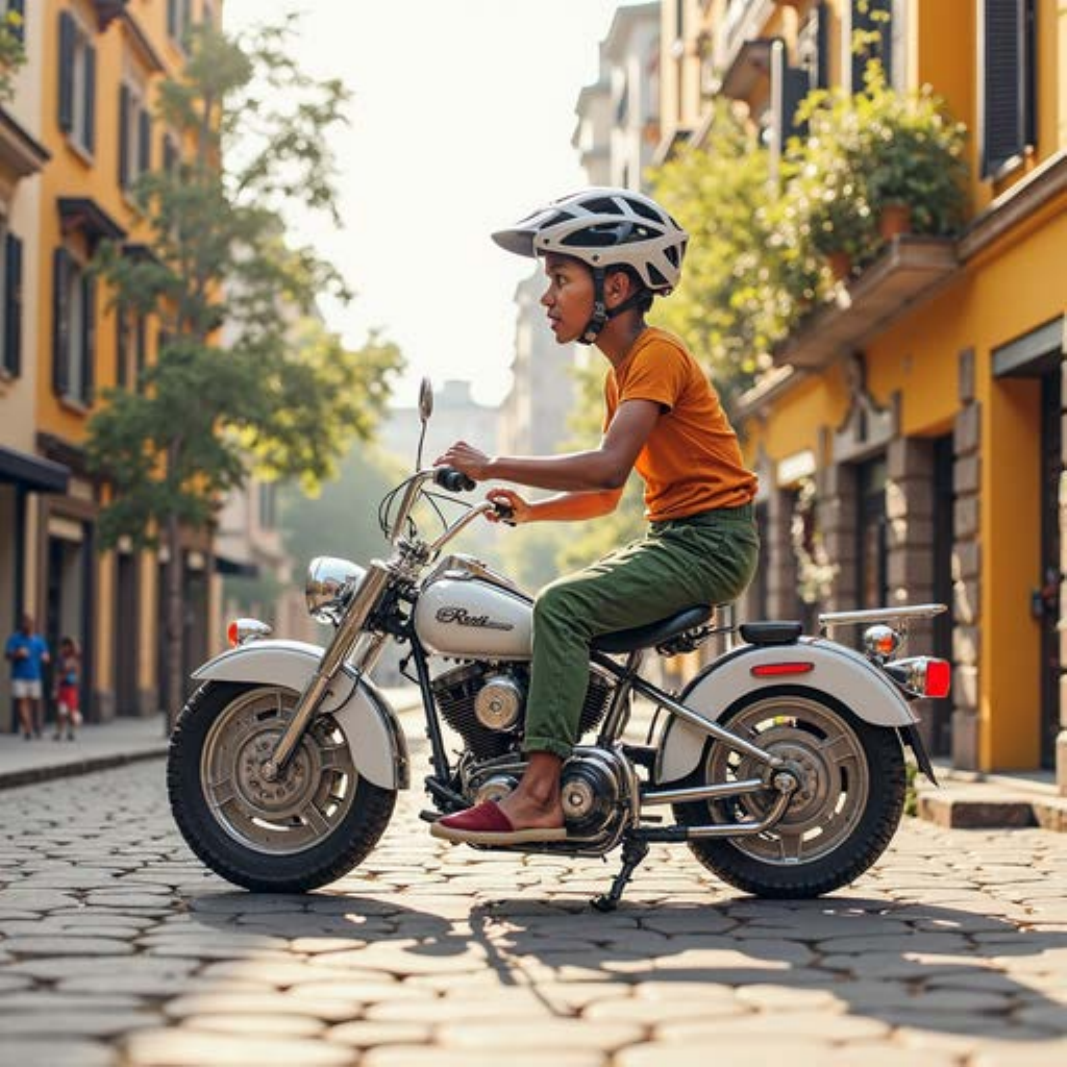}
    \end{minipage} \begin{minipage}{0.05\textwidth}
        \centering
\includegraphics[width=\textwidth]{img/arrow.pdf}
    \end{minipage} 
        \begin{minipage}{0.2\textwidth}
        \centering
        \includegraphics[width=\textwidth]{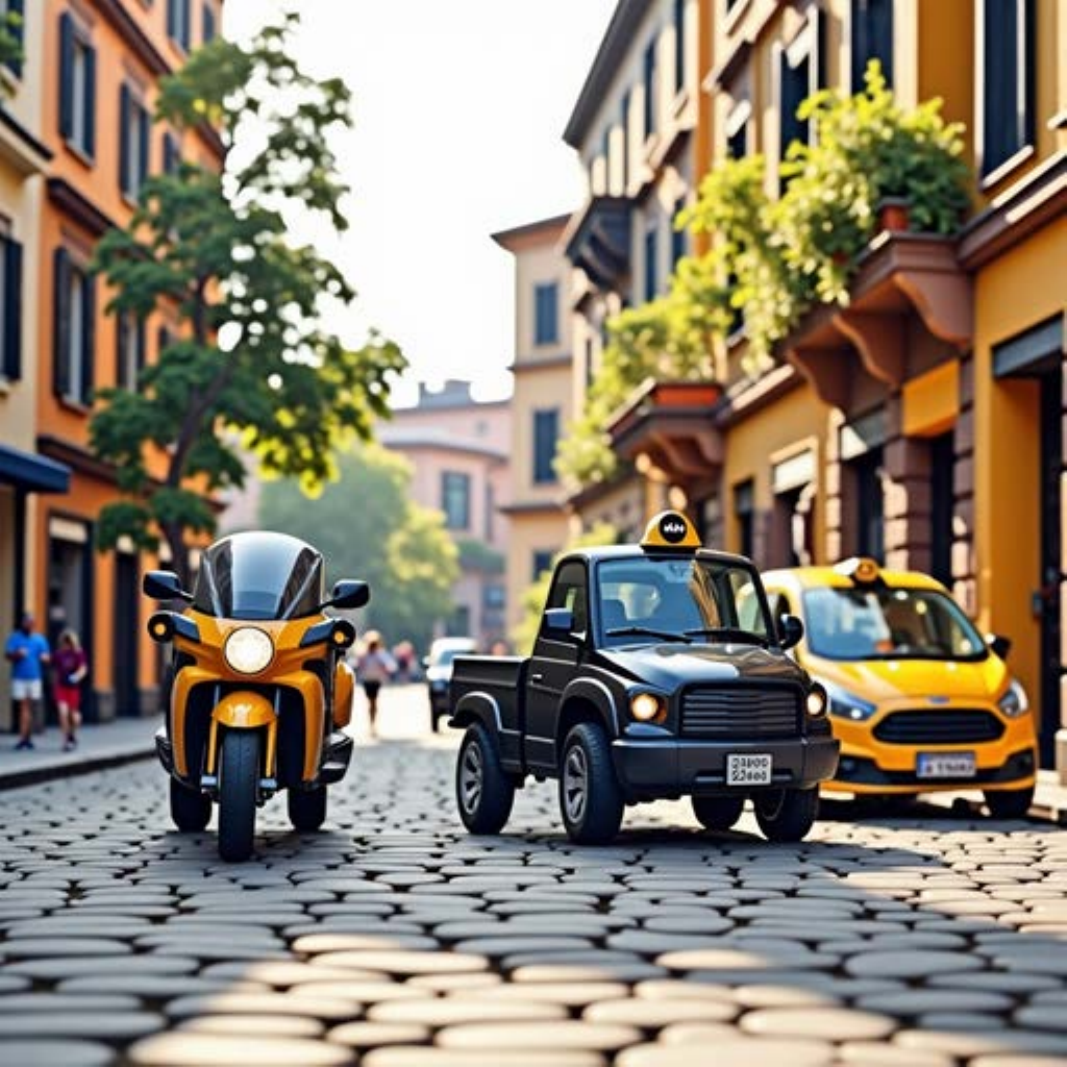}
    \end{minipage} \begin{minipage}{0.05\textwidth}
        \centering
\includegraphics[width=\textwidth]{img/arrow.pdf}
    \end{minipage} 
        \begin{minipage}{0.2\textwidth}
        \centering
\includegraphics[width=\textwidth]{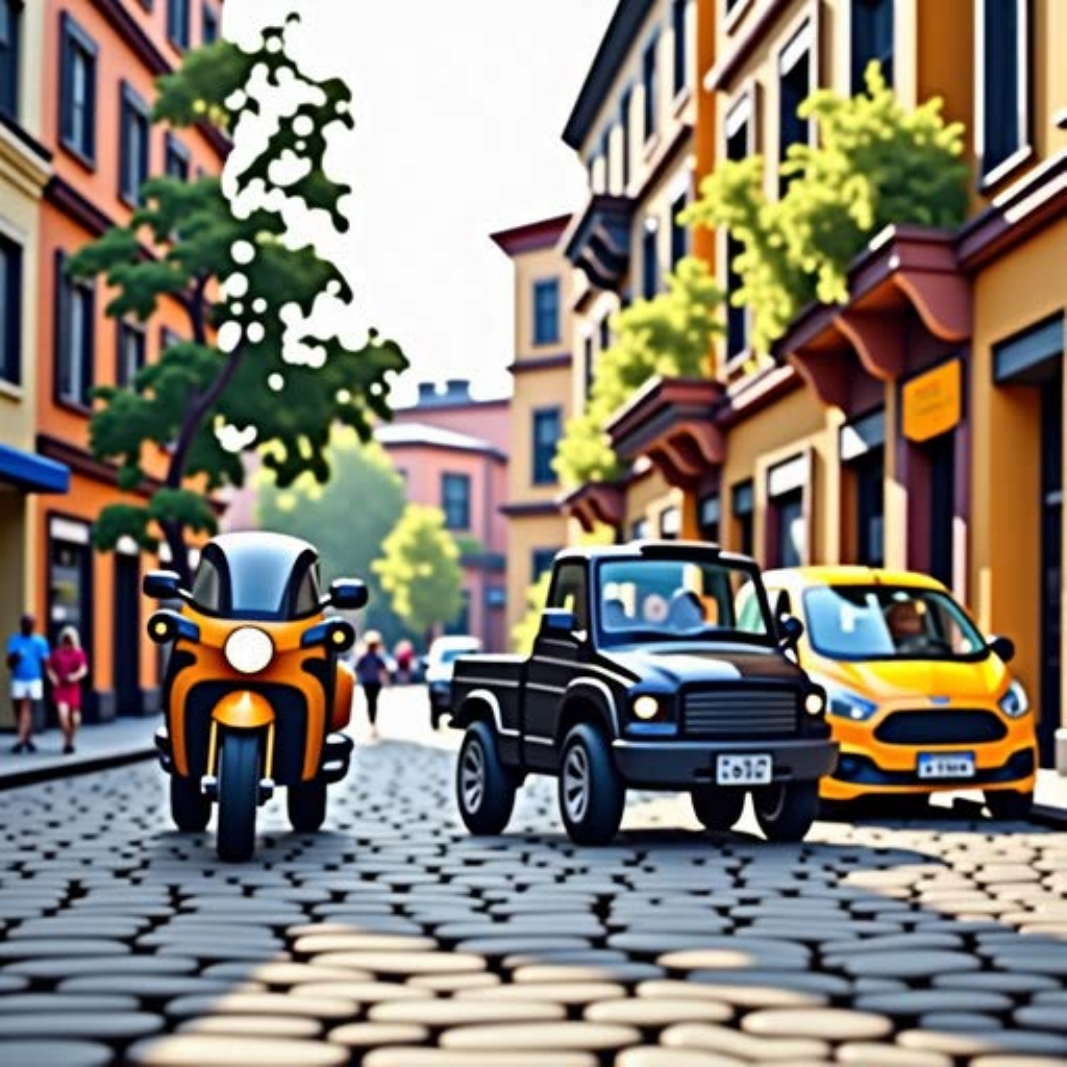}
    \end{minipage}
    \caption{A failure trial with the same prompt. In the first and second rounds, although the actor provides seemingly correct edit prompts based on feedback, the Image Editor fails to migrate old elements to the new prompts. In the third round, the Image Editor fails to make any changes, resulting in a vague image that it is unable to properly interpret or refine. From our observations, most failure cases are caused by the Image Editor's inability to handle the vague outputs it produces itself.}
    \label{case_bad}
          \vspace{-0.5em}
  \end{figure*}

In contrast, under identical conditions but with a different random seed, \autoref{case_bad} illustrates a failed attempt where the editor either completely overwrites the original image, resulting in a fundamentally different scene, or makes no changes at all. This comparison highlights that our RL-RIG design is capable of reasoning and correcting unsatisfactory outputs from the base model, whereas its performance remains partially constrained by the limitations of the base model itself.

\begin{table*}[htbp]
\centering
\caption{Comparison of more ablation studies. }
\vspace{-0.5em}
\begin{tabular}{lccccc}
\toprule
Method & SG-IoU & Ent-IoU & Rel-IoU & Qwen-Judge & GPT-Judge \\
\midrule
w/o Actor (Unsatisfied prompt) & 0.3051 & 0.8176 & 0.7150 & 0.6739 & 0.5549 \\
w/o Actor (Same prompt)        & 0.3357 & 0.9089 & 0.7632 & 0.7932 & 0.7234 \\
w/o reflection (pass @ 10)      & 0.3468 & 0.9156 & 0.7729 & 0.8392 & 0.7273 \\
\bottomrule
\end{tabular}
\label{tab:results}
\end{table*}

\section{Scene Graphs Extraction}
This part is realized by the LAION-SG dataset \cite{li2024laion}. Specifically, for each ground-truth and generated image, GPT-4o is used to output structured JSON containing scene graph triplets in ["subject", "predicate", "object"] format, along with object and predicate lists, using the following prompt:

\begin{lstlisting}
    Please extract the scene graph of the given image. 
The scene graph just needs to include the relations of the salient objects and exclude the background. 
The scene graph should be a list of triplets like ["subject", "predicate", "object"].
Both subject and object should be selected from the following list: {unique_items_list}.
The predicate should be selected from the following list: {unique_relations_list}.
Besides the scene graph, please also output the objects list in the image like ["object1", "object2", ..., "object"].
The object should be also selected from the above-mentioned object list. The output should only contain the scene graph and the object list.

Return the results in the following JSON format:
{{
    "scene_graph": [
        ["subject", "predicate", "object"],
        ...
    ],
    "object_list": [
        "object1", "object2", ...
    ],
    "predicate_list":[
        "predicate1", "predicate2", ...
    ]
}}
\end{lstlisting}

For evaluation, \texttt{{unique\_items\_list}} and \texttt{{unique\_relations\_list}} are limited only to the items that exist in the ground truth image’s scene graph. Scene graph IoU, object IoU and relationship IoUs are then calculated using the common approach.

\section{Prompt of VLM Actor and Checker}
Here we provide the prompt of our VLM Checker. 
\begin{lstlisting}
    How many of the following {length} descriptions does this image satisfy?
    -----
    {new_prompt}
    -----
    Please analyze step by step, and finally give the number of satisfied descriptions, ranging from 0 to {length}, in \\boxed{{}} format.
\end{lstlisting}

Regarding generalization, VLM Checker does have a general judgement ability, for it is substantially a generic VLM reasoning model without post training. As demonstrated in \cite{deng2025boostinggeneralizationreasoningvision}, the base VLM model performs well on multiple tasks including image QA, detection, classification and maths, which proves its generalization capabilities. 

We also provide the prompt of our VLM Actor:
\begin{lstlisting}
    Please first analyze how many descriptions this image satisfies. Then, please provide a prompt to edit the image in order to satisfy all the spatial relationships, and put the prompt in \\boxed{}. e.g., \\boxed{smiling face, long hair}. The descriptions are:
    -----
    {descriptions}
    -----
    Please analyze one by one.
\end{lstlisting}

  \section{More Ablation Studies}
  \label{sec:ablation}
To show that naive image editing approach does not work well, we add the following ablation studies.
\begin{enumerate}
    \item Not using the Actor, feed original prompt to Image Editor iteratively.
    \item Similar to 1 but only using unsatisfied constraints in the original prompt.
    \item Generate 10 images and pick the best one by VLM Checker.
\end{enumerate}

The results are shown in \autoref{tab:results}. It can be observed that prompt engineering approaches perform no better than the original Flux. We surmise that this phenomenon stems from two main factors. First, inversion-based image editing relies on guidance from the input text prompt. Repeatedly feeding the same prompt leads to identical guidance signals, resulting in little to no improvement over single-pass editing. Moreover, each round of inversion and reversion introduces additional noise, which can degrade image quality. Second, the Image Editor is highly sensitive to the style of the edit prompt. In practice, directly reusing the original prompt often fails to produce the desired modifications, as it may not align well with the editor’s preferred input style.

Although Pick 1 from 10 approach has a relatively fair performance, it is extremely computational expensive, whose inference time is more than 300 seconds per each image after pipelining, which is more than 8x times slower due to the generation cost. In fact, this result exactly fit our assumption that Reinforcement Learning is actually \emph{pruning} unnecessary generation trajectories, instead of creating new ones. As a result, RL can transfer the performance of \texttt{pass@k} to \texttt{pass@1}, but can hardly optimize \texttt{pass@1} over \texttt{pass@k}. Additionally, our RL-RIG's generator can actually be based on `Pick 1 from 10' if needed and further improve its performance.

%% file: main.bbl
\begin{thebibliography}{10}
\providecommand{\url}[1]{#1}
\csname url@samestyle\endcsname
\providecommand{\newblock}{\relax}
\providecommand{\bibinfo}[2]{#2}
\providecommand{\BIBentrySTDinterwordspacing}{\spaceskip=0pt\relax}
\providecommand{\BIBentryALTinterwordstretchfactor}{4}
\providecommand{\BIBentryALTinterwordspacing}{\spaceskip=\fontdimen2\font plus
\BIBentryALTinterwordstretchfactor\fontdimen3\font minus
  \fontdimen4\font\relax}
\providecommand{\BIBforeignlanguage}[2]{{%
\expandafter\ifx\csname l@#1\endcsname\relax
\typeout{** WARNING: IEEEtran.bst: No hyphenation pattern has been}%
\typeout{** loaded for the language `#1'. Using the pattern for}%
\typeout{** the default language instead.}%
\else
\language=\csname l@#1\endcsname
\fi
#2}}
\providecommand{\BIBdecl}{\relax}
\BIBdecl

\bibitem{stablediffusion3}
S.~AI, ``Stable diffusion 3 medium model,''
  \url{https://huggingface.co/stabilityai/stable-diffusion-3-medium}, 2025.

\bibitem{flux2024}
B.~F. Labs, ``Flux,'' \url{https://github.com/black-forest-labs/flux}, 2024.

\bibitem{deepseek2025janus}
DeepSeek, ``Janus pro: Free janus pro 7b online - ai image generator \&
  understanding,'' \url{https://januspro.io/}, 2025.

\bibitem{chen2024spatialvlm}
B.~Chen, Z.~Xu, S.~Kirmani, B.~Ichter, D.~Sadigh, L.~Guibas, and F.~Xia,
  ``Spatialvlm: Endowing vision-language models with spatial reasoning
  capabilities,'' in \emph{Proceedings of the IEEE/CVF Conference on Computer
  Vision and Pattern Recognition}, 2024, pp. 14\,455--14\,465.

\bibitem{song2024hoianimator}
W.~Song, X.~Zhang, S.~Li, Y.~Gao, A.~Hao, X.~Hou, C.~Chen, N.~Li, and H.~Qin,
  ``Hoianimator: Generating text-prompt human-object animations using novel
  perceptive diffusion models,'' in \emph{Proceedings of the IEEE/CVF
  Conference on Computer Vision and Pattern Recognition}, 2024, pp. 811--820.

\bibitem{zhang2023adding}
L.~Zhang, A.~Rao, and M.~Agrawala, ``Adding conditional control to
  text-to-image diffusion models,'' in \emph{Proceedings of the IEEE/CVF
  international conference on computer vision}, 2023, pp. 3836--3847.

\bibitem{li2023gligen}
Y.~Li, H.~Liu, Q.~Wu, F.~Mu, J.~Yang, J.~Gao, C.~Li, and Y.~J. Lee, ``Gligen:
  Open-set grounded text-to-image generation,'' in \emph{Proceedings of the
  IEEE/CVF conference on computer vision and pattern recognition}, 2023, pp.
  22\,511--22\,521.

\bibitem{lukovnikov2024layout}
D.~Lukovnikov and A.~Fischer, ``Layout-to-image generation with localized
  descriptions using controlnet with cross-attention control,'' \emph{arXiv
  preprint arXiv:2402.13404}, 2024.

\bibitem{zhao2023uni}
S.~Zhao, D.~Chen, Y.-C. Chen, J.~Bao, S.~Hao, L.~Yuan, and K.-Y.~K. Wong,
  ``Uni-controlnet: All-in-one control to text-to-image diffusion models,''
  \emph{Advances in Neural Information Processing Systems}, vol.~36, pp.
  11\,127--11\,150, 2023.

\bibitem{chen2024training}
M.~Chen, I.~Laina, and A.~Vedaldi, ``Training-free layout control with
  cross-attention guidance,'' in \emph{Proceedings of the IEEE/CVF winter
  conference on applications of computer vision}, 2024, pp. 5343--5353.

\bibitem{radford2021learning}
A.~Radford, J.~W. Kim, C.~Hallacy, A.~Ramesh, G.~Goh, S.~Agarwal, G.~Sastry,
  A.~Askell, P.~Mishkin, J.~Clark \emph{et~al.}, ``Learning transferable visual
  models from natural language supervision,'' in \emph{International conference
  on machine learning}.\hskip 1em plus 0.5em minus 0.4em\relax PmLR, 2021, pp.
  8748--8763.

\bibitem{salimans2016improved}
T.~Salimans, I.~Goodfellow, W.~Zaremba, V.~Cheung, A.~Radford, and X.~Chen,
  ``Improved techniques for training gans,'' \emph{Advances in neural
  information processing systems}, vol.~29, 2016.

\bibitem{heusel2017gans}
M.~Heusel, H.~Ramsauer, T.~Unterthiner, B.~Nessler, and S.~Hochreiter, ``Gans
  trained by a two time-scale update rule converge to a local nash
  equilibrium,'' \emph{Advances in neural information processing systems},
  vol.~30, 2017.

\bibitem{zhang2018unreasonable}
R.~Zhang, P.~Isola, A.~A. Efros, E.~Shechtman, and O.~Wang, ``The unreasonable
  effectiveness of deep features as a perceptual metric,'' in \emph{Proceedings
  of the IEEE conference on computer vision and pattern recognition}, 2018, pp.
  586--595.

\bibitem{hessel2021clipscore}
J.~Hessel, A.~Holtzman, M.~Forbes, R.~L. Bras, and Y.~Choi, ``Clipscore: A
  reference-free evaluation metric for image captioning,'' \emph{arXiv preprint
  arXiv:2104.08718}, 2021.

\bibitem{jayasumana2024rethinking}
S.~Jayasumana, S.~Ramalingam, A.~Veit, D.~Glasner, A.~Chakrabarti, and
  S.~Kumar, ``Rethinking fid: Towards a better evaluation metric for image
  generation,'' in \emph{Proceedings of the IEEE/CVF Conference on Computer
  Vision and Pattern Recognition}, 2024, pp. 9307--9315.

\bibitem{zhang2024multiview}
J.~Zhang, G.~Zhu, S.~Li, X.~Liu, H.~Song, X.~Tang, and C.~Feng, ``Multiview
  scene graph,'' \emph{Advances in Neural Information Processing Systems},
  vol.~37, pp. 17\,761--17\,788, 2024.

\bibitem{pan2025prometheus}
S.~Lee, S.~Kim, S.~Park, G.~Kim, and M.~Seo, ``Prometheus-vision:
  Vision-language model as a judge for fine-grained evaluation,'' in
  \emph{Findings of the Association for Computational Linguistics ACL 2024},
  2024, pp. 11\,286--11\,315.

\bibitem{zheng2023layoutdiffusion}
G.~Zheng, X.~Zhou, X.~Li, Z.~Qi, Y.~Shan, and X.~Li, ``Layoutdiffusion:
  Controllable diffusion model for layout-to-image generation,'' in
  \emph{Proceedings of the IEEE/CVF Conference on Computer Vision and Pattern
  Recognition}, 2023, pp. 22\,490--22\,499.

\bibitem{xue2023freestyle}
H.~Xue, Z.~Huang, Q.~Sun, L.~Song, and W.~Zhang, ``Freestyle layout-to-image
  synthesis,'' in \emph{Proceedings of the IEEE/CVF conference on computer
  vision and pattern recognition}, 2023, pp. 14\,256--14\,266.

\bibitem{wang2025stay}
R.~Wang, X.~Hou, S.~Schmedding, and M.~F. Huber, ``Stay diffusion: Styled
  layout diffusion model for diverse layout-to-image generation,'' in
  \emph{2025 IEEE/CVF Winter Conference on Applications of Computer Vision
  (WACV)}.\hskip 1em plus 0.5em minus 0.4em\relax IEEE, 2025, pp. 3855--3865.

\bibitem{farshad2023scenegenie}
A.~Farshad, Y.~Yeganeh, Y.~Chi, C.~Shen, B.~Ommer, and N.~Navab, ``Scenegenie:
  Scene graph guided diffusion models for image synthesis,'' in
  \emph{Proceedings of the IEEE/CVF International Conference on Computer
  Vision}, 2023, pp. 88--98.

\bibitem{wang2025scene}
F.~Wang, T.~Zhang, Y.~Wang, X.~Zhang, X.~Liu, and Z.~Cui, ``Scene
  graph-grounded image generation,'' \emph{Proceedings of the AAAI Conference
  on Artificial Intelligence}, vol.~39, no.~7, pp. 7646--7654, 2025.

\bibitem{black2023training}
K.~Black, M.~Janner, Y.~Du, I.~Kostrikov, and S.~Levine, ``Training diffusion
  models with reinforcement learning,'' \emph{arXiv preprint arXiv:2305.13301},
  2023.

\bibitem{wallace2024diffusion}
B.~Wallace, M.~Dang, R.~Rafailov, L.~Zhou, A.~Lou, S.~Purushwalkam, S.~Ermon,
  C.~Xiong, S.~Joty, and N.~Naik, ``Diffusion model alignment using direct
  preference optimization,'' in \emph{Proceedings of the IEEE/CVF Conference on
  Computer Vision and Pattern Recognition}, 2024, pp. 8228--8238.

\bibitem{gu2024diffusion}
Y.~Gu, Z.~Wang, Y.~Yin, Y.~Xie, and M.~Zhou, ``Diffusion-rpo: Aligning
  diffusion models through relative preference optimization,'' \emph{arXiv
  preprint arXiv:2406.06382}, 2024.

\bibitem{schulman2017proximal}
J.~Schulman, F.~Wolski, P.~Dhariwal, A.~Radford, and O.~Klimov, ``Proximal
  policy optimization algorithms,'' \emph{arXiv preprint arXiv:1707.06347},
  2017.

\bibitem{rafailov2023direct}
R.~Rafailov, A.~Sharma, E.~Mitchell, C.~D. Manning, S.~Ermon, and C.~Finn,
  ``Direct preference optimization: Your language model is secretly a reward
  model,'' \emph{Advances in Neural Information Processing Systems}, vol.~36,
  pp. 53\,728--53\,741, 2023.

\bibitem{yin2024relative}
Y.~Yin, Z.~Wang, Y.~Gu, H.~Huang, W.~Chen, and M.~Zhou, ``Relative preference
  optimization: Enhancing llm alignment through contrasting responses across
  identical and diverse prompts,'' \emph{arXiv preprint arXiv:2402.10958},
  2024.

\bibitem{ouyang2022training}
L.~Ouyang, J.~Wu, X.~Jiang, D.~Almeida, C.~Wainwright, P.~Mishkin, C.~Zhang,
  S.~Agarwal, K.~Slama, A.~Ray \emph{et~al.}, ``Training language models to
  follow instructions with human feedback,'' \emph{Advances in neural
  information processing systems}, vol.~35, pp. 27\,730--27\,744, 2022.

\bibitem{pan2025metaspatial}
Z.~Pan and H.~Liu, ``Metaspatial: Reinforcing 3d spatial reasoning in vlms for
  the metaverse,'' \emph{arXiv preprint arXiv:2503.18470}, 2025.

\bibitem{guo2025deepseek}
D.~Guo, D.~Yang, H.~Zhang, J.~Song, R.~Zhang, R.~Xu, Q.~Zhu, S.~Ma, P.~Wang,
  X.~Bi \emph{et~al.}, ``Deepseek-r1: Incentivizing reasoning capability in
  llms via reinforcement learning,'' \emph{arXiv preprint arXiv:2501.12948},
  2025.

\bibitem{wen2023improving}
S.~Wen, G.~Fang, R.~Zhang, P.~Gao, H.~Dong, and D.~Metaxas, ``Improving
  compositional text-to-image generation with large vision-language models,''
  \emph{arXiv preprint arXiv:2310.06311}, 2023.

\bibitem{li2025reflect}
S.~Li, K.~Kallidromitis, A.~Gokul, A.~Koneru, Y.~Kato, K.~Kozuka, and
  A.~Grover, ``Reflect-dit: Inference-time scaling for text-to-image diffusion
  transformers via in-context reflection,'' \emph{arXiv preprint
  arXiv:2503.12271}, 2025.

\bibitem{sun2025marmot}
J.~Sun, H.~Wang, J.~Cao, H.~Huang, and R.~He, ``Marmot: Multi-agent reasoning
  for multi-object self-correcting in improving image-text alignment,''
  \emph{arXiv preprint arXiv:2504.20054}, 2025.

\bibitem{jiang2025t2i}
D.~Jiang, Z.~Guo, R.~Zhang, Z.~Zong, H.~Li, L.~Zhuo, S.~Yan, P.-A. Heng, and
  H.~Li, ``T2i-r1: Reinforcing image generation with collaborative
  semantic-level and token-level cot,'' \emph{arXiv preprint arXiv:2505.00703},
  2025.

\bibitem{wang2024taming}
J.~Wang, J.~Pu, Z.~Qi, J.~Guo, Y.~Ma, N.~Huang, Y.~Chen, X.~Li, and Y.~Shan,
  ``Taming rectified flow for inversion and editing,'' \emph{arXiv preprint
  arXiv:2411.04746}, 2024.

\bibitem{patel2024steering}
M.~Patel, S.~Wen, D.~N. Metaxas, and Y.~Yang, ``Steering rectified flow models
  in the vector field for controlled image generation,'' \emph{arXiv preprint
  arXiv:2412.00100}, 2024.

\bibitem{rout2024semantic}
L.~Rout, Y.~Chen, N.~Ruiz, C.~Caramanis, S.~Shakkottai, and W.-S. Chu,
  ``Semantic image inversion and editing using rectified stochastic
  differential equations,'' \emph{arXiv preprint arXiv:2410.10792}, 2024.

\bibitem{Sutton2018}
R.~S. Sutton and A.~G. Barto, \emph{Reinforcement Learning: An
  Introduction}.\hskip 1em plus 0.5em minus 0.4em\relax MIT Press, 2018.

\bibitem{qwen3technicalreport}
\BIBentryALTinterwordspacing
Q.~Team, ``Qwen3 technical report,'' 2025. [Online]. Available:
  \url{https://arxiv.org/abs/2505.09388}
\BIBentrySTDinterwordspacing

\bibitem{wu2025qwenimagetechnicalreport}
\BIBentryALTinterwordspacing
C.~Wu, J.~Li, J.~Zhou, J.~Lin, K.~Gao, K.~Yan, S.~ming Yin, S.~Bai, X.~Xu,
  Y.~Chen, Y.~Chen, Z.~Tang, Z.~Zhang, Z.~Wang, A.~Yang, B.~Yu, C.~Cheng,
  D.~Liu, D.~Li, H.~Zhang, H.~Meng, H.~Wei, J.~Ni, K.~Chen, K.~Cao, L.~Peng,
  L.~Qu, M.~Wu, P.~Wang, S.~Yu, T.~Wen, W.~Feng, X.~Xu, Y.~Wang, Y.~Zhang,
  Y.~Zhu, Y.~Wu, Y.~Cai, and Z.~Liu, ``Qwen-image technical report,'' 2025.
  [Online]. Available: \url{https://arxiv.org/abs/2508.02324}
\BIBentrySTDinterwordspacing

\bibitem{shen2025vlm}
H.~Shen, P.~Liu, J.~Li, C.~Fang, Y.~Ma, J.~Liao, Q.~Shen, Z.~Zhang, K.~Zhao,
  Q.~Zhang, R.~Xu, and T.~Zhao, ``Vlm-r1: A stable and generalizable r1-style
  large vision-language model,'' \emph{arXiv preprint arXiv:2504.07615}, 2025.

\bibitem{openr1}
\BIBentryALTinterwordspacing
H.~Face, ``Open r1: A fully open reproduction of deepseek-r1,'' January 2025.
  [Online]. Available: \url{https://github.com/huggingface/open-r1}
\BIBentrySTDinterwordspacing

\bibitem{ghosh2023geneval}
D.~Ghosh, H.~Hajishirzi, and L.~Schmidt, ``Geneval: An object-focused framework
  for evaluating text-to-image alignment,'' \emph{Advances in Neural
  Information Processing Systems}, vol.~36, pp. 52\,132--52\,152, 2023.

\bibitem{li2024laion}
Z.~Li, C.~Meng, Y.~Li, L.~Yang, S.~Zhang, J.~Ma, J.~Li, G.~Yang, C.~Yang,
  Z.~Yang \emph{et~al.}, ``Laion-sg: An enhanced large-scale dataset for
  training complex image-text models with structural annotations,'' \emph{arXiv
  preprint arXiv:2412.08580}, 2024.

\bibitem{openai2024gpt4o}
OpenAI, ``Gpt-4o system card,''
  \url{https://openai.com/index/gpt-4o-system-card/}, 2024, accessed:
  2025-05-14.

\bibitem{yue2025doesreinforcementlearningreally}
\BIBentryALTinterwordspacing
Y.~Yue, Z.~Chen, R.~Lu, A.~Zhao, Z.~Wang, Y.~Yue, S.~Song, and G.~Huang, ``Does
  reinforcement learning really incentivize reasoning capacity in llms beyond
  the base model?'' 2025. [Online]. Available:
  \url{https://arxiv.org/abs/2504.13837}
\BIBentrySTDinterwordspacing

\bibitem{rafailov2024directpreferenceoptimizationlanguage}
\BIBentryALTinterwordspacing
R.~Rafailov, A.~Sharma, E.~Mitchell, S.~Ermon, C.~D. Manning, and C.~Finn,
  ``Direct preference optimization: Your language model is secretly a reward
  model,'' 2024. [Online]. Available: \url{https://arxiv.org/abs/2305.18290}
\BIBentrySTDinterwordspacing

\bibitem{wang2025implicitrewardbridgeunified}
\BIBentryALTinterwordspacing
B.~Wang, Q.~Cheng, R.~Peng, R.~Bao, P.~Li, Q.~Guo, L.~Li, Z.~Zeng, Y.~Zhou, and
  X.~Qiu, ``Implicit reward as the bridge: A unified view of sft and dpo
  connections,'' 2025. [Online]. Available:
  \url{https://arxiv.org/abs/2507.00018}
\BIBentrySTDinterwordspacing

\bibitem{li2025generalistrewardmodelsinside}
\BIBentryALTinterwordspacing
Y.-C. Li, T.~Xu, Y.~Yu, X.~Zhang, X.-H. Chen, Z.~Ling, N.~Chao, L.~Yuan, and
  Z.-H. Zhou, ``Generalist reward models: Found inside large language models,''
  2025. [Online]. Available: \url{https://arxiv.org/abs/2506.23235}
\BIBentrySTDinterwordspacing

\bibitem{zhao2025learningreasonexternalrewards}
\BIBentryALTinterwordspacing
X.~Zhao, Z.~Kang, A.~Feng, S.~Levine, and D.~Song, ``Learning to reason without
  external rewards,'' 2025. [Online]. Available:
  \url{https://arxiv.org/abs/2505.19590}
\BIBentrySTDinterwordspacing

\bibitem{yu2025rlprextrapolatingrlvrgeneral}
\BIBentryALTinterwordspacing
T.~Yu, B.~Ji, S.~Wang, S.~Yao, Z.~Wang, G.~Cui, L.~Yuan, N.~Ding, Y.~Yao,
  Z.~Liu, M.~Sun, and T.-S. Chua, ``Rlpr: Extrapolating rlvr to general domains
  without verifiers,'' 2025. [Online]. Available:
  \url{https://arxiv.org/abs/2506.18254}
\BIBentrySTDinterwordspacing

\bibitem{rafailov2024rqlanguagemodel}
\BIBentryALTinterwordspacing
R.~Rafailov, J.~Hejna, R.~Park, and C.~Finn, ``From $r$ to $q^*$: Your language
  model is secretly a q-function,'' 2024. [Online]. Available:
  \url{https://arxiv.org/abs/2404.12358}
\BIBentrySTDinterwordspacing

\bibitem{deng2025boostinggeneralizationreasoningvision}
\BIBentryALTinterwordspacing
H.~Deng, D.~Zou, R.~Ma, H.~Luo, Y.~Cao, and Y.~Kang, ``Boosting the
  generalization and reasoning of vision language models with curriculum
  reinforcement learning,'' 2025. [Online]. Available:
  \url{https://arxiv.org/abs/2503.07065}
\BIBentrySTDinterwordspacing

\end{thebibliography}
